\documentclass[a4paper, 12pt]{report}
\usepackage[utf8]{inputenc}
\usepackage[english,italian]{babel}
\pdfoutput=1
\usepackage{textcomp}
\usepackage{steinmetz}
\usepackage{url}
\usepackage{booktabs}     % Per tabelle
\usepackage{color}
\usepackage{indentfirst}  % Rientro prima riga
\usepackage{graphicx}
\usepackage{subfigure}
\usepackage{pdfpages}
\usepackage{subfigure}
\usepackage{wrapfig}
\usepackage{siunitx}      % Per unit� di misura
\usepackage{eurosym}      % Per simbolo euro
\usepackage{enumitem}     % Per la numerazione
\usepackage{algorithmic}  % Per algoritmi
\usepackage{algorithm}    % Per algoritmi
\usepackage{listings}     % Per l'ambiente lstlisting
\usepackage{pifont}       % Per elenchi particolari
\usepackage{geometry}
\usepackage{setspace}     % Per cambiare l'interlinea
\usepackage{shapepar}     % Per scrivere testi di forme particolari
\usepackage{amsmath}      % Per matrici
\usepackage{verbatim}
\usepackage{float}
\usepackage{siunitx}
\usepackage{titling}

\usepackage[printonlyused]{acronym}
\usepackage[autostyle,italian=guillemets]{csquotes}
\usepackage[backend=bibtex, style=numeric, sorting=none]{biblatex}
\usepackage{tikz}
\definecolor{mygreen}{rgb}{0,0.6,0}

\onehalfspace % interlinea 1.5

\pretitle{\begin{center}
		\begin{figure}[t]
			\centering
			\includegraphics[width=0.9\linewidth]{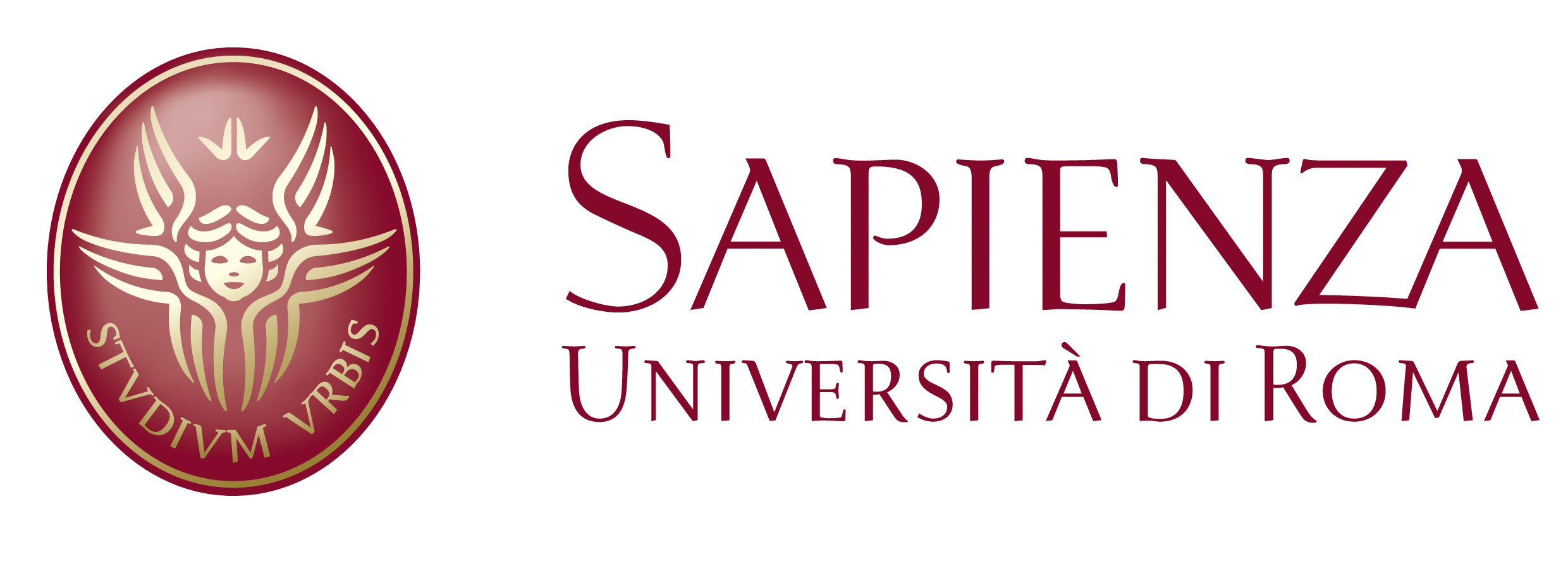}
	\end{figure}}
	
	\title{{A recurrent multi-scale approach to RBG-D Object Recognition} \\
		\bigskip }
	
	\posttitle{\textit{Candidate}: Mirco Planamente\\ \bigskip
    \textit{Thesis Advisor}: Barbara Caputo\\ \bigskip 
    \textit{Thesis External Advisor}: Mohammad R. Loghmani \\
		\bigskip 
		Facoltà di Ingegneria dell'Informazione, Informatica e Statistica\\
		Department of Computer, Control, and Management Engineering\\
		Master of Science in Artificial Intelligence and Robotics\\
		Sapienza University of Rome\end{center}}

\date{A.A. 2017/2018}
	
\begin{document}
\selectlanguage{english}
\maketitle
%----------------------------------------------------------------------------------------
%	INSERIMENTO COPERTINA
%----------------------------------------------------------------------------------------
\includepdf[pagecommand={\thispagestyle{empty}}]{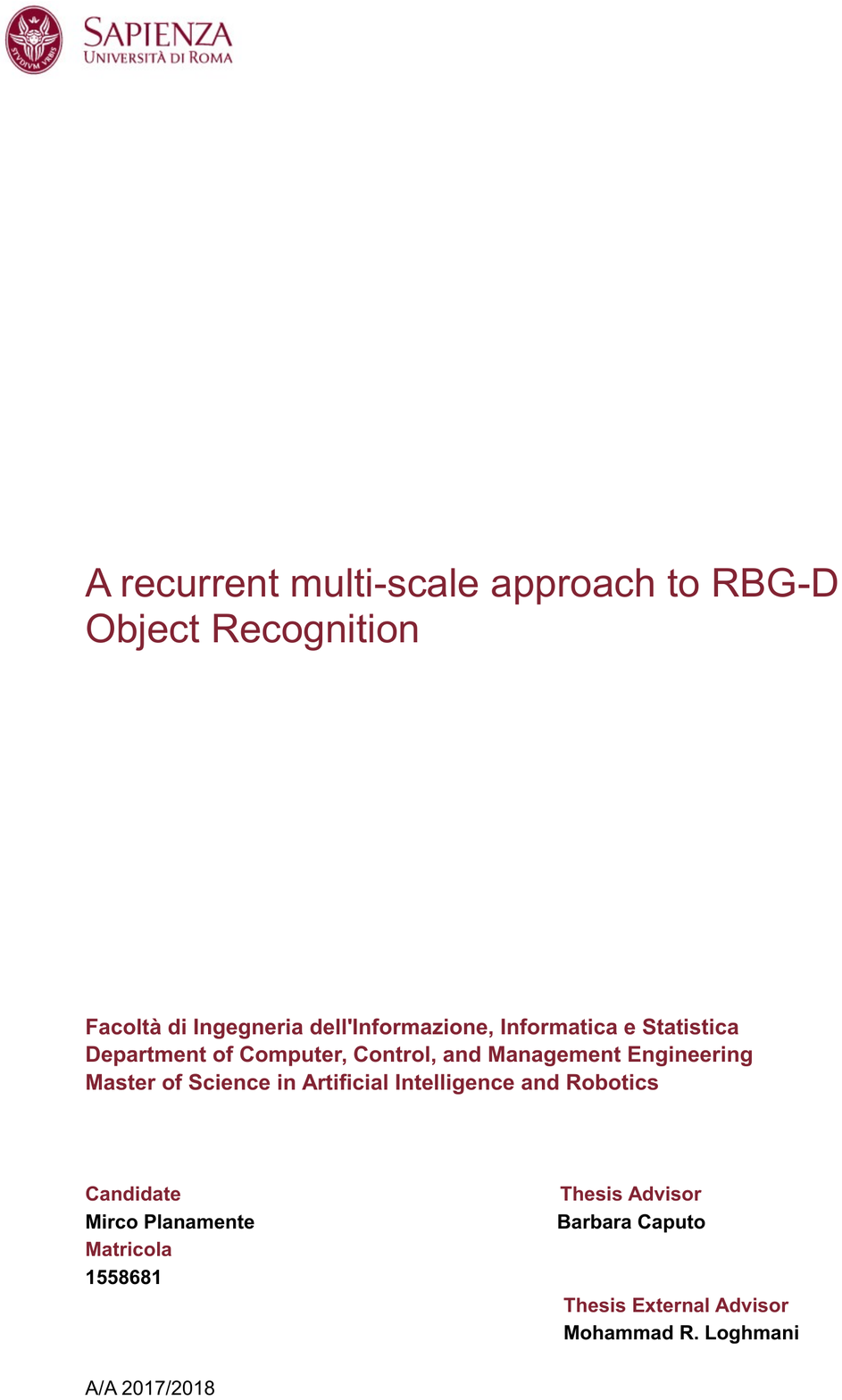}
%----------------------------------------------------------------------------------------
%	INSERIMENTO FRONTESPIZIO
%----------------------------------------------------------------------------------------

%\includepdf[pagecommand={\thispagestyle{empty}}]{Frontespizio.pdf}
%----------------------------------------------------------------------------------------
%	INSERIMENTO DEDICA
%----------------------------------------------------------------------------------------
\cleardoublepage
%\thispagestyle{empty}
%\vspace*{\fill}
\centerline{\large\bfseries Ringraziamenti}
\nobreak
%\vspace{2pc}
\begingroup\small
%\leftskip=0.1\textwidth
%\rightskip=\leftskip
\noindent

Vorrei ringraziare innanzitutto la Professoressa Barbara Caputo, relatrice di questa tesi, per avermi sempre seguito nell’arco dell’intero svolgimento di questo progetto,  concedendomi l'opportunità di svolgere parte di questo lavoro presso la Technische Universität Wien. Proseguo con il ringraziare il dottorando Mohammad R. Loghmani che mi ha dedicato molto del suo tempo e con cui ho collaborato durante lo svolgimento della tesi. A lui devo anche una mia personale crescita accademica.\\
Un ringraziamento speciale ai miei genitori che sono stati sempre presenti e vicini, anche con semplici ma importanti gesti. Sono riusciti  a sopportare i periodi in cui non ero mai in casa, per un motivo o un altro e l'ultimo periodo, dove praticamente avevo messo le radici sul divano.\\
Non potevano mancare nei miei ringraziamenti mia sorella Jessica e mio fratello Gabriele.  
Cara sister adesso per favore non confondermi con Howard Wolowitz. So benissimo che nonostante la giacca e la cravatta noi ingegneri per te resteremo sempre degli Oompa-Loompas della scienza.\\
Al mio A-ttore preferito che dire,  ormai tra la tua carriera  e i miei infortuni non abbiamo avuto molto tempo per un bel OneVsOne al campetto di casa ma sicuramente troverai altri modi per farmi del male. 
In ogni modo grazie ad entrambi, nonostante le litigate, le partite a baseball finite male e le rose delle spine, dove gentilmente da piccolo mi avete gettato, vi ringrazio per esserci sempre stati. Un ringraziamento anche a Tommaso che ormai è uno di famiglia, mi raccomando ricorda al presidente di commissione che il tuo voto può confermare o ribaltare il giudizio finale.\\

Un ringraziamento particolare è per la mia ragazza/migliore Amica/collega Silvia Bucci. Di momenti insieme ne abbiamo passati tanti ma veramente tanti, siamo stati il gossip di molte persone quì ad ingegneria. 
Molte sono le cose che mi vengono in testa e che mi piacerebbe ricordare.
I mille progetti fatti insieme, dove magari avevamo appena litigato e finiva che uno dei due se ne andava via. Le risate infinite: tipo quella volta che per sbaglio ho urtato il tuo pc e mi hai fulminato con lo sguardo ed hai iniziato a ridere fino a piangere. 
Poi tutte le ansia pre-esame, le domande sugli argomenti che non capivi, i chiusoni folli per ogni esame, il ricopiare gli appunti in bella, ad ogni tuo “ho finito” seguiva un mio “ho appena cominciato”, un tuo “come hai fatto questo esercizio?” un mio “sto studiando ancora la teoria” ed ad ogni tuo “adesso ripeto tutto per la nona volta e penso di essere pronta per domani” seguiva un mio “porca troia forse se faccio le 4 di notte in h24 riesco a fini la teoria”.\\
La tua presenza in questi anni è stata fondamentale, ti ringrazio di cuore perché so che posso contare sempre su di te.
Purtroppo non posso dilungarmi troppo con i ringraziamenti perché temo diventino più lunghi della tesi stessa, quindi ho pensato di racchiudere il nostro periodo a vienna in due parole “queue” e “happy birthday”, mi sembrano le più giuste non trovi?\\

Non potevano mancare gli amici del liceo Adrian che è stato sempre presente, le nostre chiacchierate fino alle 2/3 di notte sono epiche ormai, anche se non posso dimenticare le scale mobili a tiburtina. Martina che ti conosco da quando avevi le dita blu per l’ansia, troppe sono le risate che ci siamo fatti, soprattutto nel viaggio ad amsterdam. Luca che prima eri l'esperto di birra del gruppo e che ormai stai per diventare il nuovo pompato rubando il posto ad Adrian.
Gianluca che dopo anni continuo a chiamare Frappa, resti il nostro fotografo di fiducia. Federico, l'uomo di successo, a noi comuni mortali non ci resta che narrare i tuoi successi davanti ad una birra, ovviamente non posso non ringraziare anche rossella la nostra pr, e il nostro futuro avvocato Giada, che fanno parte del nostro gruppo da molto.\\

Ci tenevo a ringraziare tutte le persone che ho conosciuto all'università e che insieme a me hanno vissuto i periodi di studio estremo a Marta Russo come: Roberto Germanà con cui ho condiviso i migliori supplì del calabrese, nonostante il piccante non faccia per me. 
Era sempre un bel momento quello del supplì soprattutto dopo l'esame di fisica non trovi Nigga?
Come non nominare il primo molisano che ho conosciuto Andrea Scarselli, i ricordi con te sono tantissimi, sarà perché la maggior parte delle nostre serate iniziavano con una peroni 66 e finivano in situazioni epiche; come la sera che abbiamo comprato il primo cesso insieme a Germanà o quella volta dopo una serata massacrante ci siamo ritrovati a correre per strada all’Eur, non sapendo come tornare a casa.
Grazie a te ho conosciuto anche molti altri molisani, con cui ho passato belle serate e che hanno contribuito a rendere le giornate di studio meno noiose come: Cristian Chiaverini conosciuto da tutti come Chiappettini, Michele Di Pasquale, Biagio Antonelli e Luca Rossi.
Un’altra persona che voglio ringraziare è Andrea Bissoli, dopo molto tempo passato insieme e lavorando duramente sono riuscito a trasformare quel veronese sbiascicoso in un romano ad hoc. Anche con te le cose da raccontare sono molte ma penso che la più clamorosa resta la serata al Goa, ancora non mi capacito come siamo finiti in mezzo ai binari della ferrovia ma va bene così.
Non possono mancare nei miei ringraziamenti Daniele Bracciani, Ugo Nnanna Okeadu, Catenina Lacerra  e il sempre presente Atif Nura, innumerevoli sono le pause studio, caffè, pranzi e serate a San Lorenzo che ci hanno accompagnati in questi anni. 
Ci tenevo a ringraziare anche il gigante buono, il mitico Simone Agostinelli, la persone che più di tutte riesce a mettermi tranquillità, la vacanza a Gallipoli con te, Bissolino, Scarselli Fabrizio Farinacci e Daniele Zoppo resterà una delle migliori.
Un’altra persona che ci tenevo a ringraziare è Roberto Cipollone, uno dei pochi geni che ho conosciuto all'università. Il tuo metodo di studio ha influenzato moltissimo il mio, ora capita anche a me di farmi mille domande quando studio, solo che nel mio caso io non riesco a trovare le risposte che trovi tu. \\
Ci tenevo a ringraziare anche Dario Molinari con tutti i tuoi versi e le tue battute, dove solo io rido, hai reso le lezioni meno noiose. Daniele De cillis anche noto come mastro, uno dei pochi che come me sa cosa vuol dire vivere fuori dal raccordo e allo stesso tempo la persona che ha causato la scomparsa delle mollette in casa Vitali; non può mancare il nostro uomo di successo, Francesco Cucari, ti metto comunque nei ringraziamenti nonostante mi hai levato spotify craccato. Lorenzo Vitali (Pivot) finalmente qualcuno con cui ho potuto parlare di basket. Non vedo l'ora di organizzare di nuovo una bella partitella al campetto con gli altri ingegneri.
Non può mancare l'unico ingegnere che è riuscito a trovato il bug più importante, dopo il millennium bug, Stefano Carnà, ancora non ho ben capito come tu abbia trovato il trick per le macchinette, ma avrai la stima di tutti gli ingegneri a vita.
Un ringraziamento anche alla persona che è riuscito a tener in vita il mio pc, Andrea Mancini e alla miglior cuoca di ingegneria Nicoletta Spadone.
Un ringraziamento particolare anche ai ragazzi del Pegaso basket spero di potervi venire a trovare presto. I ragazzi dell'Orsa Maggiore e quelli del team Rocketz di Vienna che anche se ricordo pochissimi dei loro nomi, posso dire che quando mi dicevano “nice shot man” mi fomentavo un casino.
Un ringraziamento anche a tutti gli storici che mi hanno accolto nel loro gruppo senza discriminazioni sul fatto che studiassi ingegneria: Davide, Giovanni, Tommaso, Francesco Mattia Giulia e Simone.
Chiudo chiedendo scusa a tutti quelli che ho scordato di nominare, ma tra non molto devo consegnare la tesi e come al mio solito, da buon bradipo, sto facendo tutto all'ultimo.

\par\endgroup
\vspace{\fill}
\clearpage
\begin{abstract}
Technological development aims to produce generations of increasingly efficient robots able to perform complex tasks.
This requires considerable efforts, from the scientific community, to find new algorithms that solve computer vision problems, such as object recognition.
The diffusion of RGB-D cameras directed the study towards the research of new architectures able to exploit the RGB and Depth information.
The project that is developed in this thesis concerns the realization of a new end-to-end architecture for the recognition of RGB-D objects called \ac{RCFusion}.
Our method generates compact and highly discriminative multi-modal features by combining complementary RGB and depth information representing different levels of abstraction. 
We evaluate our method on standard object recognition datasets, RGB-D Object Dataset and JHUIT-50.
The experiments performed show that our method outperforms the existing approaches and establishes new state-of-the-art results for both datasets.

\end{abstract}
\begin{c}
This thesis is extracted from the following article~\cite{rcfusion}.
\end{c}
\tableofcontents

\listoffigures
%\addcontentsline{lof}{chapter}{List of figures}

\chapter*{List of acronyms}
%\addcontentsline{lof}{chapter}{List of acronyms}
\begin{acronym}
	\acro{SGD}{Stochastic Gradient Descent}
	\acro{CNN}{Convolutional Neural Network}
    \acro{RNN}{Recurrent Neural Network}
    \acro{FC}{Fully Connected layer}
    \acro{RCFusion}{Recurrent Convolutional Fusion}
    \acro{RelU}{Rectified Linear Units}
    
\end{acronym}

\listoftables
% Corpo--------------------------------------------------------------------
\chapter{Introduction}
\label{sec:intro}
\begin{figure}[h]
\centering
\includegraphics[width=0.428\textwidth]{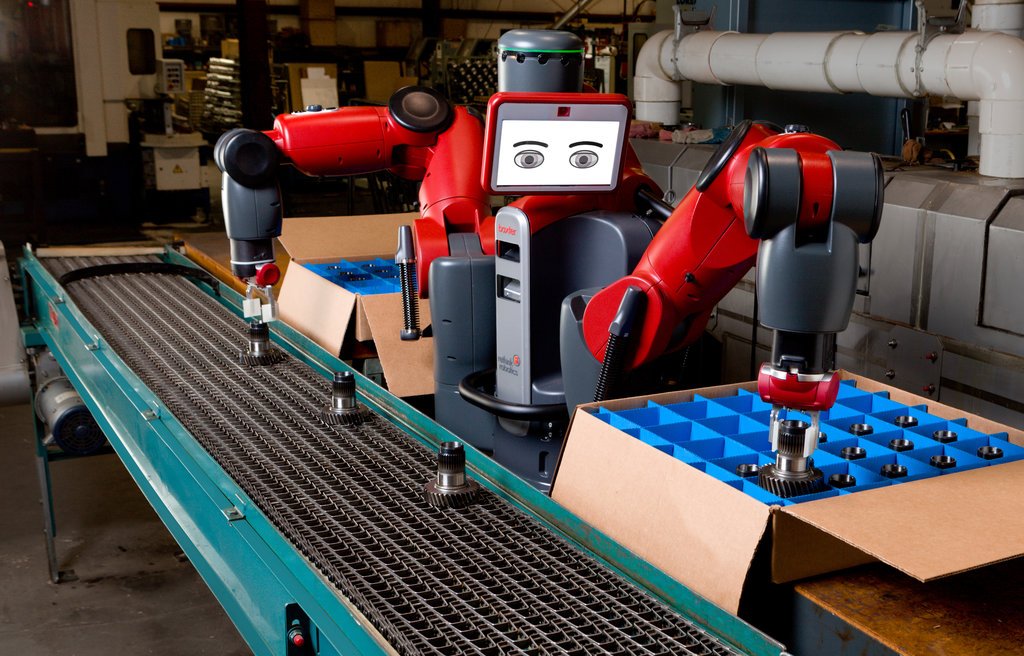}\quad \includegraphics[width=0.41\textwidth]{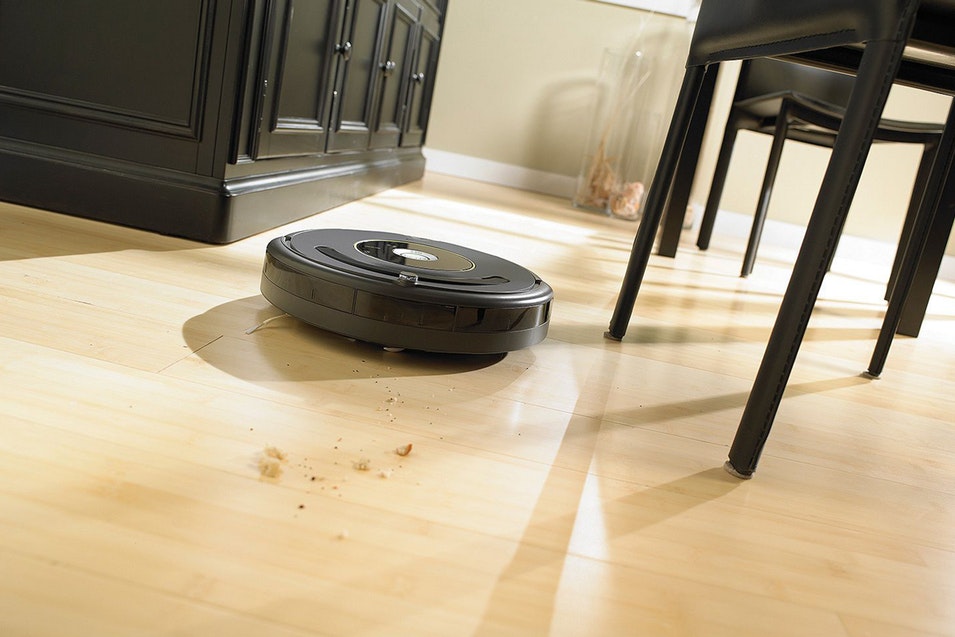}\quad \includegraphics[width=0.47\textwidth]{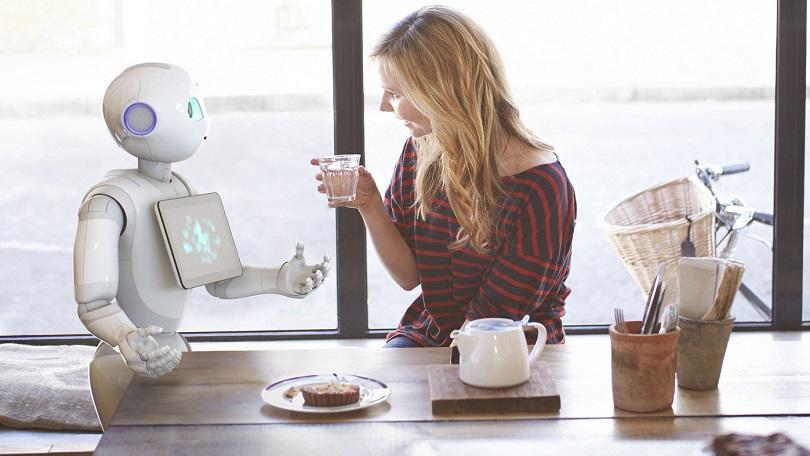}
\caption{Examples of robots that are present in human life ~\cite{robot_work},~\cite{robot_work2},~\cite{robot_work3}}
\label{fig:ex_robot}
\end{figure}

\begin{figure}[h]
\centering
\includegraphics[width=1\textwidth]{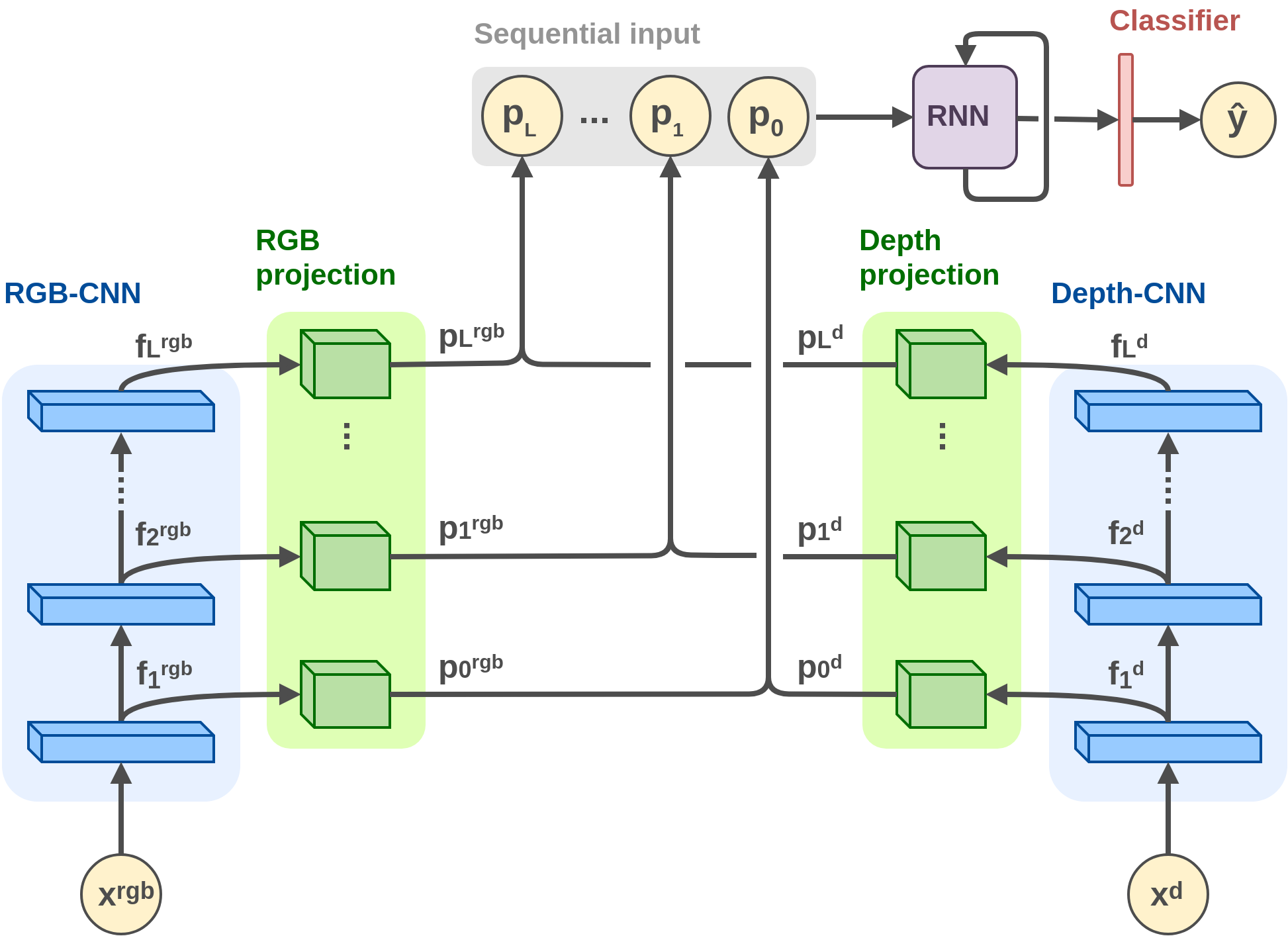} 
\caption{Reccurent Convolutional Fusion}
\label{fig:RCF}
\end{figure}

Technological progress allowed to create robots with increasingly complex capabilities, thanks to improvements both from a hardware and software point of view.
These innovations have a strong impact on human life, bringing changes in the world of work, in the life of home or in entertainment.
Figure~\ref{fig:ex_robot} shows several robots that are already present in human life and are able to perform simple tasks without the intervention of a user.
When aiming for more complex and unconstrained tasks, the ability to recognize a large variety of objects becomes a fundamental requirement for a robot operating in human environment.
One of the visual tasks that a robot must be able to solve, in order to perform high level tasks, is the object recognition.
With this term we indicate the ability to know how to find and identify the instance of an object, within an image or video.
Despite achieving great results with standard RGB images, performing object recognition with this type of data has intrinsic limitations.
In fact, the projection of the three-dimensional world into a two-dimensional image plane causes a loss of depth information.
The diffusion of low-cost RGB-D cameras allowed to add, at the RGB image, the information on the distance from the scene to the camera. It is possible to obtain better performances, using the information on the color, texture and appearance of the RGB image and the geometric information, obtained from the depth image.\\
The introduction of Convolutional Neural Networks (CNNs)  produced considerable progress in various areas of artificial intelligence research, becoming quickly the dominant tool in computer vision. CNNs allowed the achievement of new state-of-the-art results for a large variety of tasks, such as human pose estimation~\cite{openpose}, semantic segmentation~\cite{mask_rcnn} and image super-resolution~\cite{super-res}. Research in RGB-D object recognition followed the same trend, with numerous algorithms (e.g.~\cite{eitel2015,aakerberg2017,depthnet,deco,li2015,wang2015}) exploiting features learned from CNNs instead of the traditional hand-crafted features.
The basic idea is to use CNNs as features extractor for both modalities.
It is proven that the use of this type of networks allows to obtain remarkable performances when trained with very large datasets ~\cite{he2016deep}.
The lack of a data set of such dimensions introduced a limit, not allowing to exploit the data depth with the same architectures of the RGB data.\\
The scientific community has developed an approach that avoids this limitation.
The idea is to colorize depth images in order to exploit CNNs pre-trained on RGB images.
Compared to the considerable efforts to find increasingly effective colorization methods, strategies to extract and combine the characteristics of the two modalities are neglected.
In fact, several methods simply rely on a trivial concatenation of the features extracted from a specific layer of the two CNNs that are then fed to a classifier.
We believe that this approach does not allow to fully exploit the potential of CNNs because (a) it is based on the assumption that the selected layer always represents the best abstraction level to combine RGB and depth information and (b) no explicit mechanism ensures that the diverse characteristics of the two modalities are properly exploited.\\ 
The project that is developed in this thesis concerns the realization of a new end-to-end architecture for the recognition of RGB-D objects called \ac{RCFusion}.
Our method extracts features from multiple hidden layers of the CNNs for RGB and depth, respectively, and combines them through a recurrent neural network (RNN), as shown in figure~\ref{fig:RCF}. 
In addition, we formulate a loss function to promote orthogonality between corresponding RGB and depth features, assisting the two streams in learning complementary information.
In order to show the innovative aspects of this new architecture, we present different experiments with different modalities.
With a first set of experiments, we show that the combination of complementary features representing different levels of abstraction, produces a highly discriminative description of the RGB-D data.
We evaluate our method on standard object recognition benchmarks, RGB-D Object Dataset~\cite{wrgbd} and JHUIT-50~\cite{jhuit50}, and we compare the results with the best performing methods in the literature.
The experimental results show that our method outperforms the existing approaches and establishes new state-of-the-art results for both datasets.\\
In the remaining part of the thesis we show other approaches found in the literature that solve the same problem of object recognition (chapter ~\ref{chapter:related_works}), highlighting the points in common and not with our network.
In chapter ~\ref{chapter:back}, we provide the basic concepts that are related to the Neural Networks, \ac{CNN}s and at the end \ac{RNN}s, in order to have a basic knowledge that allows a better understanding the work performed.
A description of our method is given in chapter ~\ref{chapter:rcf}, thus providing an overview of the structure.
Then, in chapter ~\ref{chapter:implementation}, we provide details related to implementation.
In the last two chapters,~\ref{chapter:ex} and ~\ref{chapter:conclusion}, we show and discuss the results obtained from the network and finally, summarize the fundamental points of this work, with its possible future developments.

\chapter{Related Works}
\label{chapter:related_works}
Object recognition is one of the main problems in computer vision and important for making robots useful in home environments.
The introduction of RGB-D cameras made it possible to record color and depth images.
The development of this type of sensors directed research towards new methods and approaches to manage and exploit this new information.
Among the countless computer vision tasks that can benefit from the additional geometric information (e.g. ~\cite{gupta2014}, ~\cite{action_recog},~\cite{scene_recog}), the RGB-D object recognition task plays a fundamental role since it acts as a proxy for higher level tasks.\\
Classical approaches for RGB-D object recognition (e.g. ~\cite{bo2011}, ~\cite{wrgbd}) used a combination of different hand-crafted feature descriptors, such as SIFT ~\cite{sift}, textons ~\cite{textons}, and depth edges, on the two modalities (RGB and depth) to perform object matching. 
More recently, several methods have exploited shallow learning techniques to generate features from RGB-D data in an unsupervised learning framework ~\cite{blum2012}, ~\cite{hmp2011}, ~\cite{socher2012}. These methods rely on algorithms such as hierarchical matching pursuit ~\cite{hmp2011}, k-means clustering ~\cite{kmeans} and RNNs ~\cite{lstm} to progressively build higher level features in an unsupervised fashion.\\
The introduction of deep learning has had a strong impact on research, leading to an increase in results in the field of object recognition.
The use of CNNs as features extractors has proven to be more efficient than other methods ~\cite{decaf}, ~\cite{razavian2014}. They are able to learn filters with different levels of abstraction ranging from simpler elements such as lines or angles (in the initial parts of the network) to more complex objects (in the deeper layers).
This happens when these networks are trained (in a supervised way) with large datasets. While largescale datasets of RGB images, such as ImageNet ~\cite{imagenet}, allowed the generation of powerful CNN-based models for RGB feature extraction, the lack of a depth counterpart posed the problem of how to extract features from depth images.\\
An idea to avoid this problem is to color the depth images in order to exploit the CNN pre-trained with RGB data.
Gupta et al. ~\cite{gupta2014} proposed a kind of colorization called HHA.
At each pixel of the depth image is applied a transformation that maps the image in three channels, representing horizontal disparity, height above the ground and the angle the pixel’s local surface normal makes with the inferred gravity direction. 
This approach encodes properties of geocentric pose that emphasize complementary discontinuities in the image (depth, surface normal and height).
Eitel et al.~\cite{eitel2015} used in their project, in addition to HHA, also a simpler type of coloring called colorjet. It consists in mapping the lowest depth values in blue, the higher values in red and the intermediate values accordingly.
Demonstrating how this type of colorization can provide an inexpensive yet effective colorization.
Bo et al, ~\cite{bo2013} uses a type of colorization, that exploits surface normal, adding an image processing that aims to reconstruct the missing data of the image before coloring it.
After the processing, they calculated the normal vector to the surface for each pixel and used the values of the x, y and z components as the three channels of the image.
This approach produced slightly better results than the previous two methods.
Carlucci et al ~\cite{deco} propose a different approach. They developed a network, based on the residual paradigm, which learns how to color images depth. Producing results that are visually better and at the same time competitive with other methods.\\
Other methods try to manage the problem in a different ways.
Instead of using a colorization method, Li et al. ~\cite{li2015}  generated the depth features using a modified version of HONV~\cite{honv} encoded as Fisher vector ~\cite{fisher}.
Carlucci et al. ~\cite{depthnet} generated artificial depth data using 3D CAD models to train a CNN that extracts features from raw depth images.\\
The aforementioned methods focused on effectively extracting features from the depth data and used trivial strategies to combine the two modalities for the final prediction. 
Very few works prioritized this latter aspect. 
Wang et al. ~\cite{wang2015} alternatively maximize the discriminative characteristics of each modality and the correlation between the final features of RGB and depth data. 
Wang et al. ~\cite{cimdl} generate a multi-modal feature that explicitly separates the individual and correlated information of RGB and depth data that are then used to
predict the final label through adaptive weight softmax regression. Both these methods optimize the fusion of the RGB and depth modality on features extracted from one layer of the
CNNs selected a priori. 
The focus of this work is on the synthesis of multi-modal features from RGB-D data rather than the depth processing. In fact, for the depth processing part, we adopt the well known colorization method based on surface normals, since it has been proved to be effective and robust across tasks~\cite{bo2013,cimdl,aakerberg2017}. \\
Differently from existing works, our method does not rely on features extracted from one specific layer of a CNN, but combines features extracted from multiple layers to generate the final multi-modal representation. In addition, our model can be trained in all together, without the need of optimizing in multiple stages.

\chapter{Background}
\label{chapter:back}
In this chapter, we introduce the technical foundation of our work.
We provide the basic concepts that are related to the Neural Networks, Convolutional Neural Networks (CNNs) and at the end Recurrent Neural Networks (RNNs), in order to have a basic knowledge that allows a better understanding of the project performed in this thesis.

\section{Neural Networks}
\begin{figure}[h]
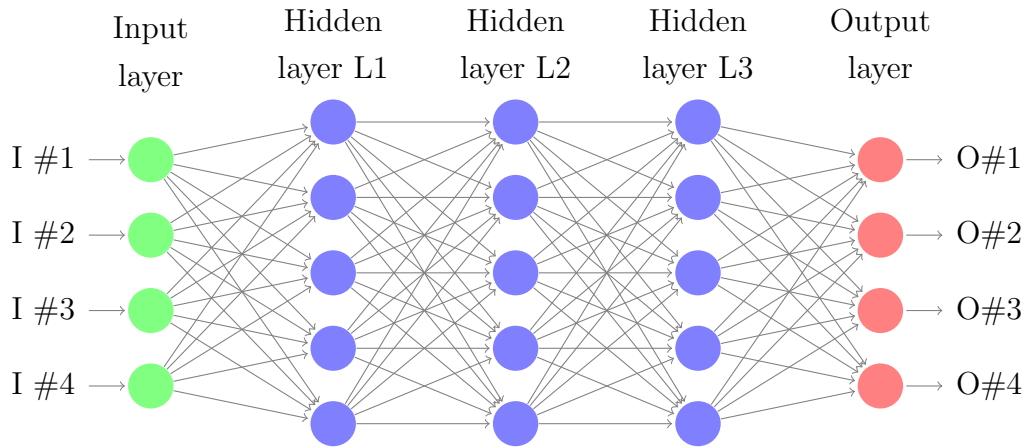

\centering
\caption{Diagram of a neural network with three hidden layers and multiple outputs.}
\label{fig:exNN}
\end{figure}

Neural networks are models of machine learning, which try to imitate the biological brain in structure and functioning. They are made of a number of basic computational elements, called neurons, interconnected with each other and grouped in order to divide the structure into layers.
Each neuron contains a function, called activation function. It consists of a threshold or limitation function that makes sure that only signals with values compatible with the threshold or the imposed limit can propagate to the next neuron or neurons.  Typically, the activation function is a nonlinear function, it is usually a step function, a sigmoid or a logistic function.
\begin{align}
  y = f( \sum^{}_{i} w_ix_i + b )
  \label{eq:funct_act}
\end{align}
As we can see from the formula, the output is dependent on the input \textit{x$_i$}  which represents the input fed to the network, in the case of the first layer or the output of a previous layer. \textit{f} represents the activation function and the other parameters \textit{w$_i$} and \textit{b} are respectively the weights assigned to each input and a bias term that is added at the end. These are the values that each neuron stores and learns during training.
For simplicity we indicate the set of parameters to be trained with capital letters \textbf{W} and 
\textbf{B}.
As you can see from the figure ~\ref{fig:exNN} the network is divided into layers and each layer is formed by different neurons. The neurons of a layer are all connected with the neurons of the previous and following layers. It is necessary to keep in mind that with increasing network depth (ie adding hidden layers) the number of parameters to be trained increases considerably.\\

\def\layersep{2.4cm}
\begin{tikzpicture}[shorten >=1pt,->,draw=black!50, node distance=\layersep]
    \tikzstyle{every pin edge}=[<-,shorten <=1pt]
    \tikzstyle{neuron}=[circle,fill=black!25,minimum size=17pt,inner sep=0pt]
    \tikzstyle{input neuron}=[neuron, fill=green!50];
    \tikzstyle{output neuron}=[neuron, fill=red!50];
    \tikzstyle{hidden neuron1}=[neuron, fill=blue!50];
    \tikzstyle{hidden neuron2}=[neuron, fill=blue!50];
    \tikzstyle{hidden neuron3}=[neuron, fill=blue!50];
    \tikzstyle{annot} = [text width=4em, text centered]

    % Draw the input layer nodes
    \foreach \name / \y in {1,...,4}
    % This is the same as writing \foreach \name / \y in {1/1,2/2,3/3,4/4}
        \node[input neuron, pin=left:I \#\y] (I-\name) at (0,-\y) {};

    % Draw the hidden layer nodes
    \foreach \name / \y in {1,...,5}
        \path[yshift=0.5cm]
            node[hidden neuron1] (H1-\name) at (\layersep,-\y cm) {};
    \foreach \name / \y in {1,...,5}
        \path[xshift=\layersep,yshift=0.5cm]
            node[hidden neuron2] (H2-\name) at (\layersep,-\y cm) {};
    \foreach \name / \y in {1,...,5}
        \path[xshift=2*\layersep,yshift=0.5cm]
            node[hidden neuron3] (H3-\name) at (\layersep,-\y cm) {};

    % Draw the output layer node
     \foreach \name / \y in {1,...,4}
   		%\node[output neuron, pin={[pin edge={->}]right:O} \#\y](O-\name) at (4*\layersep,-\y cm){};
   \node[output neuron,pin={[pin edge={->}]right:O\#\y}, right of=H3-3] (O-\name)at (3*\layersep,-\y cm) {};

    % Connect every node in the input layer with every node in the
    % hidden layer.
    \foreach \source in {1,...,4}
        \foreach \dest in {1,...,5}
            \path (I-\source) edge (H1-\dest);
    \foreach \source in {1,...,5}
        \foreach \dest in {1,...,5}
            \path (H1-\source) edge (H2-\dest);
    \foreach \source in {1,...,5}
        \foreach \dest in {1,...,5}
            \path (H2-\source) edge (H3-\dest);
            
    % Connect every node in the hidden layer with the output layer
    \foreach \source in {1,...,5}
    	\foreach \dest in {1,...,4}
       		 \path (H3-\source) edge (O-\dest);
	
    % Annotate the layers
    \node[annot,above of=H1-1, node distance=1cm] (hl) {Hidden layer L1};
    \node[annot,above of=H2-1, node distance=1cm] (hl) {Hidden layer L2};
    \node[annot,above of=H3-1, node distance=1cm] (hl) {Hidden layer L3};
    \node[annot, above of= I-2] {Input layer};
    \node[annot,right of=hl] {Output layer};

\end{tikzpicture}

\subsection{Training}

The term "learning" refers to the policy used to update the weights of the neural network in order to reduce the difference between the output produced by the network and the desired output.\\
A first phase is called \textbf{forward propagation}, in this phase the input crosses the entire network from the input layer to the output one. Before the input flows through the network for the first time the weights of each neuron have been randomized initialized with values close to zero.
Through a loss function we evaluate how far the output produced by the desired one is. After calculating the loss value, another phase begins, which takes the name of \textbf{back-propagation}. 
In this phase the error is propagated backwards (in the opposite direction with respect to the previous phase). Subsequently the weights of each neuron of the network are updated following a specific policy, and then start again with the first phase. The most used policy for updating weights is called \textbf{gradient-descent}.
The second phase starts from the last level of the network. For each neuron an error term is calculated that provides a measure of how much that neuron is responsible for the error in the network output.
The following error terms are then used to calculate the gradients of the cost function \textit{C}, which usually occurs in this form: \begin{align}
  C(W,b, x\textsuperscript{(i)} , y\textsuperscript{(i)}) = {1 \over 2} \left \| f\textsubscript{W,b}  (x\textsuperscript{(i)}) - y \textsuperscript{(i)} \right \|
  \label{eq:funct_act}
\end{align}
with respect to \textbf{W} and \textit{b}. Recall that \textit{x} represents the input of the network and \textit{y} the desired output.
The neuron weights are updated with a single gradient-descent iteration with the following update rules:\begin{align}
  W \textsuperscript{l}\textsubscript{i,j} = W \textsuperscript{l}\textsubscript{i,j} - \boldsymbol{\alpha} { \partial  \over \partial  W \textsuperscript{l}\textsubscript{i,j} } C(W,b)
  \label{eq:funct_act}
\end{align}

\begin{align}
  b \textsubscript{i}\textsuperscript{(l)} = b \textsubscript{i}\textsuperscript{(l)} - \boldsymbol{\alpha} { \partial  \over \partial  b \textsubscript{i}\textsuperscript{(l)}} C(W,b)
  \label{eq:funct_act}
\end{align}
the $\boldsymbol{\alpha}$ parameter in the formula is used to specify the degree of learning (update), with respect to the parameters \textbf{W} and \textit{b} and takes the name of the \textbf{learning rate}.
Usually this term is variable and generally decreases as training progresses. In this way, the network is able to perform larger steps in the direction of the gradient at the beginning of the training, when it is generally far from the minimum value. The entities of the updates are reduced when it begins to get close to it, so as to facilitate the convergence and avoid unwanted divergences.
A possible negative scenario deriving from the use of this technique occurs when the search for the minimum remains blocked at a local minimum, prohibiting the network to achieve the expected result.

\subsection{Stochastic Gradient Descent (SGD)}

The variations of the gradient descent algorithm may be different, which differ in the amount of data processed by the network for the calculation of the gradient, before updating the parameters. A first variation calculates the gradient only after processing the entire dataset and is called \textbf{Vanilla Gradient Descent} or \textbf{Batch Gradient Descent}.
It is proven that it is more efficient to compute the gradient after processing of a small amounts of data (batch) at a time ~\cite{ruder2016overview}. The amount of data that is processed before upgrading to network weights is called \textbf{Batch size}. We will see how it plays an important role among the parameters that influence the training and performance of the network. Generally the batch selection happens randomly, hence the name \textbf{\ac{SGD}}.
The choice of batch sizes smaller than the whole dataset is sometimes obligatory, because often the dataset's dimensions are such that they can not be loaded into memory.
Even the opposite choice, batch size too small for example equal to 1, is not a good choice.
It would lead to a reduction of the risk to remain blocked  in a local minimum but also an increase of the risk of divergence or non-convergence. As a consequence, intermediate values are usually chosen, typically in the range [32, 256].
Regardless of the size of the selected batch, every time that all \textit{n} observations of the dataset are processed by the network it is said that an \textbf{epoch} is completed; in the case of Vanilla Gradient Descent, the batch size is n, then at each iteration corresponds to an epoch.\\
The gradient descent algorithm can be modified to improve its performance, trying to avoid some problems (such as the local minimum).
Near the areas where the surface has a much more pronounced curvature in one direction than the other, the algorithm presents some difficulties.
In this scenario, in fact, the algorithm is led to oscillate along the steepest slide, thus slowing the convergence towards the optimal local minimum. \textbf{Momentum} is a method that helps to accelerate the descent of the gradient towards the correct direction, dampening the effect induced by the oscillations.
This result is obtained by adding a fraction \textit{y} of the previous update vector to the current update term.
Typically, values of \textit{y} are used around 0.9.
\begin{figure}[h]
\centering
\includegraphics[width=0.99\textwidth]{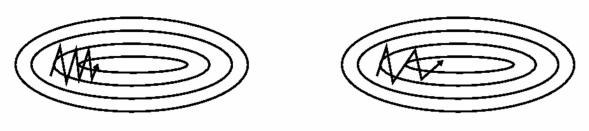}
\caption{(a) Stochastic Gradient Descent without momentum  \qquad      (b) Stochastic Gradient Descent with momentum~\cite{ruder2016overview}.}
\label{fig:momentum}
\end{figure}
In the training of models with many parameters, such as neural networks, it is necessary to use regularization forms in the estimation phase, if you do not want to have \textbf{overfitting} problems. A form of regularization that is often used is the \textbf{dropout}. With this technique some neurons are left out during the estimation phase, this happens in a completely random way. We define $\boldsymbol{\varphi}$ as the probability with which the nodes of the network between the various layers are \textit{"switched off"}. This probability can also have different values between various layers of the network (typically higher values are used in the denser and/or final layers, leaving the input layer intact).
\begin{figure}[h]
\centering
\includegraphics[width=0.869\textwidth]{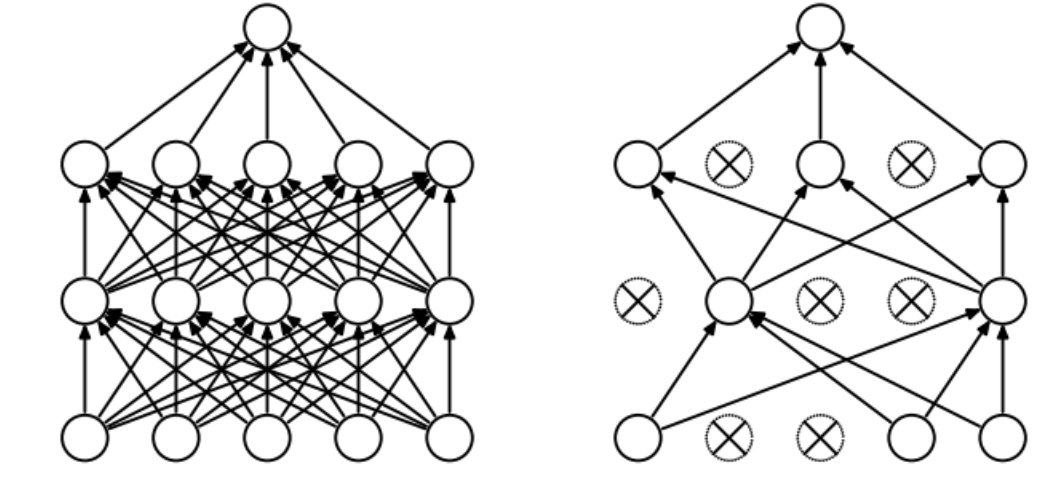}
\caption{(a) Standard Neural Network (b) After applying a dropout~\cite{srivastava2014dropout}}
\label{fig:SNN+dropout}
\end{figure}
The application of the dropout therefore produces at each iteration a different reduced network of the starting model, composed of those nodes that have survived the (temporary) elimination process. As a result, a combination of models is created that produces, in most cases, a performance improvement and a reduction in generalization error.
Obviously in the testing phase, a single complete neural network is used, whose weights are the scaled versions of the previously calculated weights:  if a neuron had a probability $\boldsymbol{\varphi}$ to be eliminated from the model during the estimation phase, then its weights will be multiplied by $\boldsymbol{\varphi}$.

\section{Convolutional Neural Networks (CNNs)}

The Convolutional Neural Networks are networks in which the connection between neurons pattern are inspired by the structure of the visual cortex in the animal world, for convenience are often called by their acronym CNN.
The single neurons present in this part of the brain respond to precise stimuli in a narrow region of the observation, called \textbf{receptive field}. The receptive fields of different neurons are partially overlapped so that they cover the entire field of view altogether. The response of a single neuron to stimuli taking place in its receptive field can be mathematically approximated by a convolution operation.
In this type of networks the neurons in the layers are organized in \textbf{width}, \textbf{height} and \textbf{depth}, like in figure \ref{fig:SNN+dropout}.\\
\begin{figure}[h]
\centering
\includegraphics[width=0.9\textwidth]{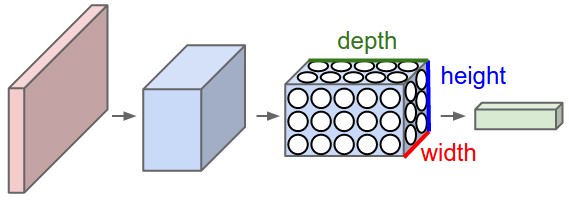} 
\caption{ A Convolutional Neural Networks~\cite{CNN_ex} }
\label{fig:SNN+dropout}
\end{figure}There is a major difference with the neural networks described above. In CNNs, the neurons of a layer are connected only to a small region of the previous layer, rather than to all the neurons. This goes to solve the problem with the scalability that the fully connected architectures have.
A classical neural network that implements convolution operations in at least one of its layers is called Convolutional.
The layers composed of convolution operations take the name of Convolutional Layers, but are not the only layers that compose a CNN: the typical architecture provides for the alternation of \textbf{Convolutional Layers}, \textbf{Pooling Layers} and \textbf{Fully Connected Layers}.

\subsection{Convolution operation}

As previously mentioned in the convolution networks each neuron is connected only to a local region of the input volume. The spatial extension of this connectivity is a hyper-parameter called the receptive field of the neuron (it is equivalent to the size of the filter). The extension of the connectivity along the depth axis is always equal to the depth of the input volume. It is important to emphasize again this asymmetry in the way we treat the spatial dimensions (width and height) and the depth dimension. There are three parameters that control the size of the output volume: \textbf{depth}, \textbf{stride} and \textbf{zero padding}.\\
The depth of the output volume is a hyper-parameter that substantially controls the number of neurons in the convolutional layers that are connected to the same input local region.
In particular, a set of neurons of this type is called, for obvious reasons, \textbf{depth column}. It corresponds to the number of filters that are used, each of which is learning to look for something different in the input.\\
The Stride is the step in which the filter translates on the input. A stride of 1 means that there will be a depth column of neurons each one spatial unit of distance from the local area of application of another depth column. This could lead to an intense overlapping of the regions and also to a considerable increase in the volume of output. The use instead of a major stride can lead to a reduction of overlapping between the regions and therefore to spatially smaller volumes. Of course all this always depends on the size of each local region, that is the receptive field (the stride rarely assumes greater values than 2).
While the filter flows along the input, the values of the scalar product between the selected data portion and the filter weights are added together. What we get is called the \textbf{features map}.
\begin{figure}[h]
\centering
\includegraphics[width=1.25\textwidth]{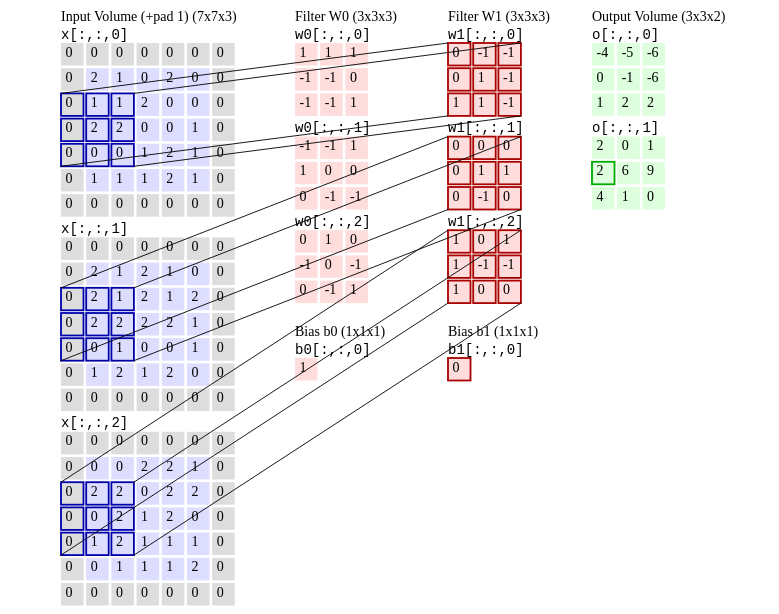} 
\caption{Convolution operation of two different filters (W0 and W1) of dimension 3×3×3 on a 7×7×3 input volume~\cite{CNN_ex}.}
\label{fig:op_conv}
\end{figure}
This will produce smaller volumes of output spatially. It is often convenient to fill the input volume with zero around the edge. The size of this zero padding is a hyper-parameter. The formula that binds the values of dimensions of the input, the receptive field and the stride is:
\begin{align}
{W - F + 2P \over S} + 1
  \label{eq:output_formula}
\end{align}
where:
\begin{itemize}
\item W is the width of the input,
\item F is the dimension of the receptive field,
\item P is the zero-padding parameter,
\item S is the stride parameter.
\end{itemize}
This formula returns the value of the width as a result. To obtain the height value, it is sufficient to replace the W value of the width with the height relative to the input. The interesting feature of padding zero is that it allows us to control the spatial size of the output volumes, allowing us to preserve exactly the spatial dimension of the input volume so that the input and output width and height are the same.

\subsection{Sharing parameters}

To reduce the number of parameters a reasonable assumption is made: if a given feature is useful in a position (x1, y1) then it is also useful in another position (x2, y2). From a more practical point of view, let's call depth slice a single two-dimensional "section" along the depth axis of the output volume (for example a volume X x Y x Z has Z depth slice of X x Y dimensions). All neurons in each \textbf{depth slice} use the same weight and bias. Specifically, during the backward propagation phase each neuron will calculate the gradient of its weight, but these gradients will be added through each depth slice and will update only a single set of weight for each depth slice. It is usual to refer to the weight sets with the term \textbf{filters} or, more commonly, \textbf{kernels} and with each of them a convolution operation is performed with the input. The result of this convolution is an \textbf{activation map}. The assumption made with the technique of sharing the parameters is relatively reasonable: if the identification, for example, of a horizontal line is important in a certain location of the image, it should theoretically be important also in any other location of the image.  It is therefore not necessary to learn again to identify a horizontal line in each of the different locations of the volume of output.\\
The capacity of a CNN can vary depending on the number of layers it has. A CNN can have multiple layers of the same type. In a few cases, for example, there is only one convolutional layer, unless the network in question is extremely simple. 
Usually a CNN has a series of convolutional layers: the first of these are used to obtain low-level features, such as horizontal or vertical lines, angles, various contours, etc. Continuing into the network, going towards the output layer, the features become high-level, in other words they represent more complex figures such as faces, specific objects, a scene, etc. Basically therefore, more convolutional layers have a network and more detailed features it can process. 

\subsection{Pooling Layer}

Another type of layer indispensable in a CNN is the pooling layer. These layers are periodically inserted within a network to reduce the spatial size (width and height) of the current representations, this allows to reduce the number of parameters, the computational time of the network and also keeps under control overfitting.  
A pooling layer operates on each depth slice of the input volume independently, going to resize it spatially. For resizing, a simple function is used, such as a MAX or AVE operation (the maximum or average of a given set of elements). \\
The pooling layers have some settable parameters:
\begin{itemize}
\item  the F side of the spatial extension of the square selection that will be considered, from time to time, on the input in each of its depth slices,
\item  the stride parameter S.
\end{itemize}
You can see a certain similarity with the parameters of a convolutional layer: this is because also here there is a kind of receptive field that is moved from time to time, with a step specified by the stride parameter, on each depth slice of the input volume. Actually, besides this, the situation in this case is completely different, since here there is no convolution operation.
However, just as in a convolutional layer, also here there is a relationship between the parameters of the spatial extension F and the stride S (no zero-padding), since it is still necessary to cover the entire area of each depth slice. The relation is in fact very similar to the one observed previously: considering the width W of an input volume, the output volume of a pooling layer will have width 
\begin{align}
{W- F  \over S} + 1
  \label{eq:output_formula}
\end{align}
A pooling layer has some settable parameters, but it does not have any trainable parameters, unlike a convolutional layer: in fact, once the parameters have been set, a fixed function is applied.
For this reason, a pooling layer performs its tasks only in the forward propagation phases, while in the backward propagation phases it is limited to propagating the errors.
\newpage
\begin{figure}[H]
\centering
\includegraphics[width=0.90\textwidth]{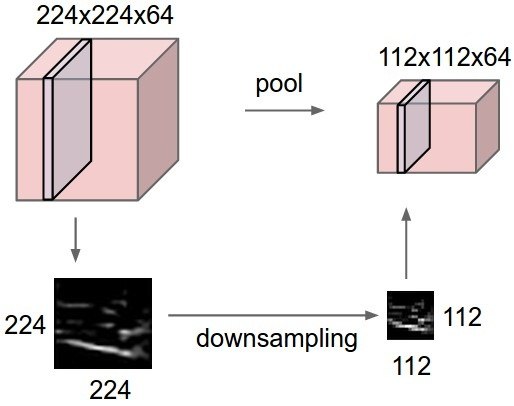} 
\label{fig:pool_1}
\end{figure}
\begin{figure}[h]
\centering
\includegraphics[width=0.97\textwidth]{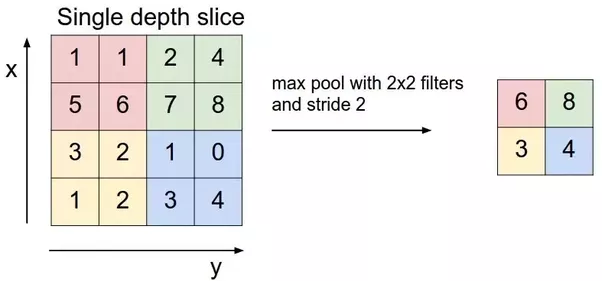} 
\caption{Effects of a general pooling operation~\cite{CNN_ex}.}
\label{fig:pool_2}
\end{figure}

\subsection{Rectified Linear Units (\ac{RelU})}

First of all, ReLU is the abbreviation of Rectified Linear Units. This type of layer is very common in a CNN and is used several times within the same network, very often after each convolutional layer. Its main function is to increase the non-linearity property of the activation function without going to modify the receptive fields of a convolutional layer. A function very suitable for this purpose is 
\begin{align}
f(x)= \max(0,x)
  \label{eq:output_formula}
\end{align}
However, you can specify other layer types with the same purpose but which use different functions. The ReLU layers allow you to train a network very quickly, this explains why they are very used. As you can guess, this type of layer has no parameter that can be set or train, simply a fixed function is performed. Layers not trainable parameters, as we have seen, we have a backward propagation simpler: the calculated errors are retro-propagate until then, coming from the next layer, passing the previous layer.

\subsection{Fully Connected}

This type of layer is exactly the same as any of the layers of a classical artificial neural network with fully connected architecture: simply in a \ac{FC}, each neuron is connected to all the neurons of the previous layer, specifically to their activations. Therefore the activation of a neuron of a FC layer can be calculated with the product between the weight matrix and the input matrix followed by the addition of a bias. \\
This type of layer, unlike what has been seen so far in CNNs, does not use the local connectivity property: a FC layer is connected to the entire input volume and therefore there will be many connections. The only settable parameter of this type of layer is the number of neurons K that it has. What basically makes a FC layer is to connect its K neurons with all the input volume and to calculate the activation of each of its K neurons. Its output will be a single 1x1xK vector, containing the calculated activations. After using a single FC layer, the input volume, organized in 3 dimensions, becomes a single output vector, in a single dimension, this suggests that after the application of an FC layer, no convolutional layer can be used.\\
The main function of FC layers in the context of convolutional neural networks is to group the information obtained, expressing it with a single number (the activation of one of its neurons), which will be used in subsequent calculations for the final classification. Usually more than one FC layer is used in series and the last of these will have the parameter K equal to the number of classes present in the dataset. The final K values will be finally fed to the output layer, which, through a specific probabilistic function, will perform the classification. Also in this type of layer there are some trainable parameters (the weight and the bias) and then in the backward propagation through the usual backpropagation algorithm and with the gradient-descent method they are updated.

\subsection{Residual networks}

The research community is making a lot of effort to develop innovative deep architectures. We can see how the levels of the networks increase, starting from AlexNet ~\cite{krizhevsky2012imagenet} a CNN architecture formed by 5 convolutional levels passing VGG ~\cite{vgg} and GoogleNet ~\cite{szegedy2015going} which have respectively 19 and 22 levels.
However, increasing depth always leads to problems. Too deep networks are difficult to train because of the known problem of the vanishing of gradient. 
Because it is propagated back to the previous levels and repeated multiplications can make the gradient infinitely small.  As a result, as the network goes deeper, its performance gets saturated or even starts degrading rapidly. 
There are several ways to handle this problem, add an auxiliary loss in an intermediate level as extra supervision ~\cite{szegedy2015going}, but this does not seem to really solve the problem. \\
% * <mirco.pl.93@gmail.com> 2018-05-02T15:15:31.853Z:
% lo strovi nel sito salvato trai preferiti 
% ^.
A solution to this problem can be seen in Resnet ~\cite{he2016deep} with the introduction of a new mechanism called \textbf{residual block}.
The concept on which the residual block is based is to give an input \textit{x} to the
sequence of convolution-ReLU-convolution operations, obtaining a certain
 \textit{F(x)}, and to add the same \textit{x} to the result.
As a result of this block, we have  \textit{H(x)} =  \textit{F(x)} + \textit{x} . In a traditional feed forward CNN, in practice, it would have instead that  \textit{H(x)} =  \textit{F(x)}.
Through the concatenation of different blocks of this type, ResNet learns to predict a certain output, not through the learning of a direct transformation from the input data to the output, but through the learning of a certain term  \textit{F(x)}.
This term is to be added to the input data to arrive at the output minimizing the error, which is called residual error. This approach is called \textbf{residual learning}.\\
\begin{figure}[h]
\centering
\includegraphics[width=0.95\textwidth]{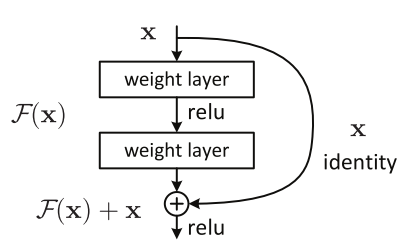} 
\caption{A residual block~\cite{resnet}.}
\label{fig:Residual}
\end{figure}
Another component that is widely used in ResNet is represented by the levels of \textbf{batch normalization} ~\cite{ioffe2015batch}, used after each convolution and activation. Batch normalization is an operation that allows to normalize the data present in the mini-batches, and thanks to this it reduces the limitations on the value of the learning rate that typically exists in the training of deep neural networks. It also makes the initialization phase of the weights less complex. All this leads to a considerable reduction in the time required for the network training process.
The central idea in the ResNet paper is that, in building a neural network with a high number of levels. The representation of input data should remain as unaltered as possible going deep into the network, in order to preserve information.\\
This architecture won ILSVRC 2015 with an error of 3.6.\%.  To understand the value of this result, just think that the error generally achieved by a human is around 5-10\%, based on his skills and knowledge.

\section{Recurrent Neural Networks}

Recurrent networks are a type of artificial neural network designed to process sequential data. With this particular structure we introduce another analogy with the human brain linked to the concept of Memory. The networks presented until now are based on input (and output) independent of each other. However, these networks present difficulties in some tasks that require the network to remember the data processed in the past. \\
The RNNs with their particular structure are able to memorize in their hidden layers a state vector that implicitly contains information on the history of all the elements of the sequence's past. We can see this capacity as a sort of network memory.\\
Another feature of the RNNs is the idea of sharing parameters in different parts of the model. This property makes it possible to extend and apply the model to examples of different forms of data, increasing the ability to generalize the network (thinking of speech recognition in which data can be letters, words or phrases of different sizes).	
\begin{figure}[h]
\centering
\includegraphics[width=1.1\textwidth]{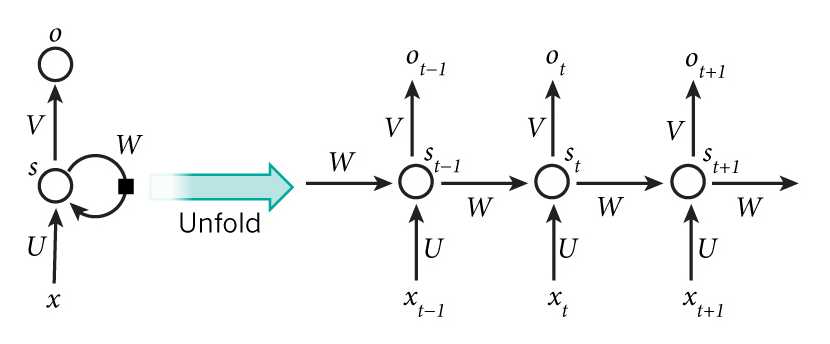} 
\caption{A recurrent neural network and the unfolding in time of the computation involved in its forward computation~\cite{RNN_1}.}
\label{fig:RNN}
\end{figure}
Considering the output of the hidden layers at different times of the sequence as outputs of different neurons of a deep multi-layer neural network, it becomes easy to apply backward propagation to train the network.\\ However, although the RNNs are powerful dynamic systems, the training phase is often very problematic because the gradient obtained with backward propagation or increases or decreases at any discrete time, so after many moments of time it can become too large or not considerable. 
Hidden units grouped in node \textit{s} with value \textit{s$_i$} take the input from the neurons of the previous phase. In this way the network can map an input sequence formed by the \textit{x$_t$} elements into an output sequence of the \textit{o$_t$} elements, where each \textit{o$_t$} depends on all the input data \textit{x$_{t0-t}$} at instants prior to t (t0 $<$ t).
The same parameters (the matrices U, V, W) are re-used at each next phase.\\

\subsection{Training of Recurrent Neural Networks}

The training of an RNN is very similar to that of a traditional neural network. The backpropagation algorithm is always present but with a small variation. Since the parameters are shared by all the time phases of the network, the gradient of each output depends on the calculations of the current time step and also on the previous time steps.
This type of backpropagation is called \textbf{Backpropagation Through Time (BPTT)}. The algorithm is applied directly to the computational graph obtained by deploying the sequential branch of the network. In this case, the network can be considered as a multi-layer network of which each layer represents a single cycle of the sequence, having shared weights.\\ 
The architecture of the RNNs makes them very efficient in solving the tasks in the field of speech and language recognition as the prediction of the next character, or word, that will be found in a text. Despite the main goals of recurring networks in long-term learning, theoretical and empirical evidence shows that it is difficult to learn by storing much information for very long time sequences.
Indeed, the network often focuses on recent information. Data learned at much earlier times may result in errors during training.\\
The use of the RNN produces a good result in cases where the gap between the relevant information and the context of the sentence is small. However, there are situations that introduce difficulties using the RNNs with the structure presented until now.
Suppose the case of the sentence  \textit{"I grew up in Italy ... I speak a \textbf{***} sliding"}, recent information suggests that the right word is probably \textit{"Italian"}, but nothing prevents the word from being another. 
In this case the gap between relevant information and the point at which the sentence must be completed is very large. The few relevant information, do not forbid the word is another language as \textit{"French"}, \textit{"Spanish"}, etc.
In this type of situation the RNNs have poor performance to learn the correct information. The solution to this problem is to increase the size of the network by adding explicit memory.\\
One type of networks that implement this solution are \textbf{Long Short-Term Memory networks (LSTM)}. These networks use special hidden layers formed by units specialized in remembering input for very large time intervals.
A special unit called a memory cell acts as an accumulator, as if the network's neuron was provided with a permeable membrane (gate): it has connections on itself to the next time with a unit weight, so it can copy the real value of the state by accumulating the external signals; this auto-connection is connected to an educated unit to decide when to delete the contents of the memory.\\
All recurring networks are in the form of a chain of neural network modules that are repeated. In standard recurrent networks the repetition module has a very simple structure, like a single layer with activation function the hyperbolic tangent (tanh), often preferred to other activation functions because the calculation of the gradient is less expensive.
The concatenated structure of LSTMs allows instead to have a single layer with multiple interacting neurons following this particular scheme.
\begin{figure}[h]
\centering
\includegraphics[width=1.1\textwidth]{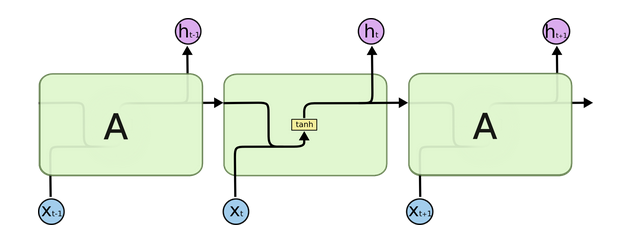} 
\caption{ A traditional recurrent neural network with the repetition module has one
very simple structure given by a single layer having a hyperbolic tangent unit~\cite{RNN_2}.}
\label{fig:RNN_chain}
\end{figure}
The state cell is represented by the line in the figure that flows over the whole chain; a few linear operations are applied to them, allowing the information to travel unaltered.
The network has the ability to remove or add information to the status cell through the gate structure. These structures are composed of a neural layer with a sigmoidal activation function and a punctual multiplication operation. 
The output values of this state, between 0 and 1, quantify how much information from the input must be allowed to flow in the network: a value of 0 means that nothing should pass while the value 1 indicates the total passage of information.\\
An LSTM has three gates to protect and control the status cell.
The first, called forget gate layer, decides which information you want to enter in the network flow; the second, input gate layer, decides which values must be updated, immediately after a hyperbolic tangent layer creates a vector having as elements the new candidate values to be added to the state; finally, the last gate decides which part of the state vector must be returned in output.
\begin{figure}[h]
\centering
\includegraphics[width=0.95\textwidth]{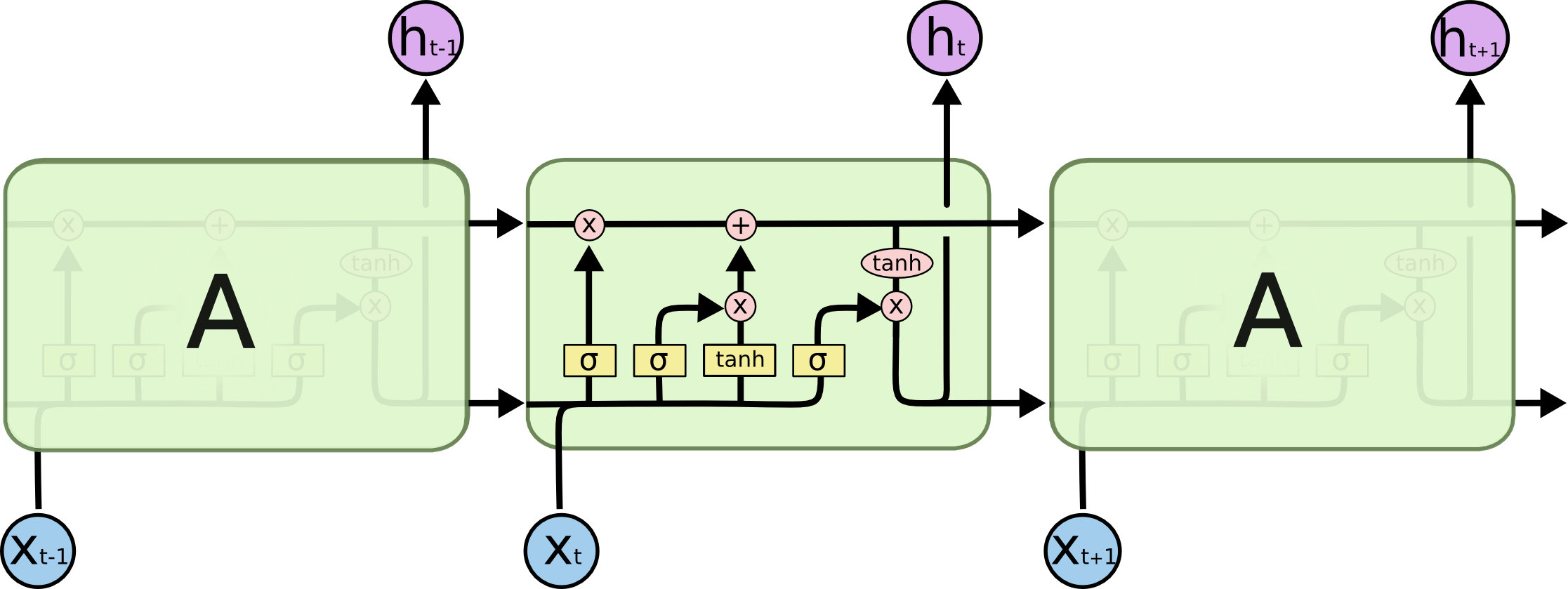} 
\caption{ Long Short-Term Memory networks in which the three gates are present~\cite{RNN_2}.}
\label{fig:LSTM_chain}
\end{figure}
It has been proven that in many applications LSTMs are more effective than traditional RNNs, especially when they have many layers at each moment of time. These networks achieve good performance in a speech recognition system that acts from the acoustic to the transcription of the character.
The gate units of the LSTMs are also used for encoding and decoding networks which perform the translation task well.

\chapter{Recurrent Convolutional Fusion}
\label{chapter:rcf}
The structure of the network that we have proposed is illustrated in the figure ~\ref{fig:RCF}.
It is a multi-modal deep neural network for RGB-D object recognition.
Our method acts simultaneously on features extracted from multiple layers and imposes a soft orthogonality constraint to extract complementary information from the two modalities.\\
Three are the main features of this network:\begin{enumerate}
\item  \textbf{multi-level feature extraction}: The data  RGB and Depth are processed using two CNNs. The architecture of the CNNs used is the same for both modality. We extract the features at different levels of the networks (RGB-Cnn and Depth-Cnn);   
\item  \textbf{feature projection and concatenation}: We apply a transformation through projection blocks at each features. This features are after concatenated in order to create the correspondent RGB-D feature;
\item \textbf{recurrent multi-modal fusion} RGB-D features extracted from different levels are sequentially fed to a recurrent network that produces a more descriptive and compact multi-modal feature.
\end{enumerate}
The CNN that we have choice for both modality is Resnet18.
The train of the network can be made end-to-end using standard backpropagation algorithms based on stochastic gradient descent.
The objective function that we want minimize is composed by two part.
The first is the standard classification loss and the second is an \textbf{orthogonality loss}.
We describe better and with more details all this parts of the network that we have just mention.

\section{Multi-level feature extraction}

The first novelty of our model is in the extraction of the features at different level.
As described above, during image processing, CNNs filters learn features with different levels of abstraction: from simple lines or angles for lower levels, to objects and scenes in higher levels.
By observing different methods found in literature it is possible to find a basic idea among them.
Most of them exploit the RGB-D data, going to use the output of one of the last layers of the used CNNs, to process respectively the two modality.
This is because they are based on the strong assumption that the chosen layer represents the appropriate level of abstraction to combine the two modalities. With our model we want to introduce a “stronger” assumption than the one just described. We want to show that by exploiting also the other levels of abstraction of the features and combining them in an appropriate way it is possible to obtain a highly discriminative RGB-D feature.\\
Let us denote with $x^{rgb} \in \mathcal{X}^{rgb}$ the RGB input images, with $x^{d} \in \mathcal{X}^{d}$ the depth input images and $y \in \mathcal{Y}$ the labels, where $\mathcal{X}^{rgb}$, $\mathcal{X}^{d}$ and $\mathcal{Y}$ are the RGB/depth input and label space. We further denote with $f^{rgb}_i$ and $f^{d}_i$ the output of layer $i$ of RGB-CNN and Depth-CNN, respectively, with $i=1,...,L$ and $L$ the total number of layers of each CNN. As previously mentioned, the RGB- and Depth-CNN are assumed to have the same architecture. 
Instead of extracting the features from a layer $i$, selected a priori, we synthesize the final multi-modal representation from the output of multiple layers.\\

\section{Feature projection and concatenation}

The features extracted from the rgb and depth parts are called with the same superscripts $*$ to make writing and reading the formulas easier.
In general, the features extracted from different levels have different dimensions between them and thus belong to distinct features space $\mathcal{F}_i$ and $\mathcal{F}_j$. 
This means that $f^{*}_i$ and $f^{*}_j$, with $i \neq j$.
In order to make features coming from different levels of abstraction comparable, we project them into a common space $\bar{\mathcal{F}}$, using a projection function $G(.)$:\begin{equation}
\setlength\abovedisplayskip{5pt}
\setlength\belowdisplayskip{5pt}
  p^{*}_i = G^{*}_i(f^{*}_i) \qquad \text{s.t.} \quad
  p^{*}_i \in \bar{\mathcal{F}} \quad \forall i.
  \label{eq:risk_}
\end{equation}
The result  $p^{*}_i$ is the representation of the features $f^{*}_i$ in the common space $\bar{\mathcal{F}}$. 
This function G(.) performs a set of non-linear operations defined by pooling and convolutional layers.
The projected RGB and depth features of each layer $i$ are then concatenated to form $p_i = \big[ p^{rgb}_i ; p^{d}_i \big]$.

\section{Recurrent multi-modal fusion}

The set of concatenated features $\big\{ p_1, \dots , p_{L} \big\}$  are given sequentially in input to the RNN.In order to create a compact multi-modal representation. As the last step the output of the RNN is used, by the softmax classifier, for the final prediction $\hat{y}$.
The choice of a fully connected layer, is  usually more straightforward choice, is not possible. This is because we do not want the network parameters to be $L$ dependent and grow linearly with it.
In fact, using the RNN the parameters to be towed are independent of $L$.\\
Another reason that led us to choose this type of network is that the unit of memory of the network, if properly trained, incorporates a summary of the most relevant information from the different levels of abstraction

\section{Loss}

During the training of the entire network the weights are updated with the backward propagation algorithm. In this phase we want to minimize the loss function, with respect to the network parameters.
This function is described by the following formula:\begin{equation}
\setlength\abovedisplayskip{5pt}
\setlength\belowdisplayskip{5pt}
  \mathcal{L} = \mathcal{L_{\text{cls}}} + \mathcal{L_{\bot}}
  \label{eq:loss}
\end{equation}
The first term of equation~\ref{eq:loss} is the classification loss $\mathcal{L}_{\text{cls}}$ that trains the model to predict the output labels we are ultimately interested in. It is defined as a cross-entropy loss \begin{equation}
\setlength\abovedisplayskip{5pt}
\setlength\belowdisplayskip{5pt}
  \mathcal{L_{\text{cls}}} = - \sum^{S}_{j=1} y_j \log \hat{y}_j,
  \label{eq:loss_cls}
\end{equation}
where $S$ is the number of available training samples. We propose to regularize the training by adding an orthogonality loss $\mathcal{L_{\bot}}$ that imposes a soft orthogonality constraint between the projected RGB and depth features. Let $P^{rgb}_i$ and $P^{d}_i$ be matrices whose rows are the projected features $p^{rgb}_i$ and $p^{d}_i$. The orthogonality loss $\mathcal{L_{\bot}}$ is defined as \begin{equation}
\setlength\abovedisplayskip{5pt}
\setlength\belowdisplayskip{5pt}
  \mathcal{L_{\bot}} = \sum^{L}_{i=1} \lambda_i \left \| {{P^{rgb}_i}^T P^{d}_i }\right \|_F^2,
  \label{eq:loss_perp}
\end{equation} 
where $\lambda_i$ denotes the magnitude (/weight) of the regularization for layer $i$, $\left \|{.}\right \|_F^2$ is the squared Frobenius norm, and the superscript $T$ denotes the transpose operator. The purpose of $\mathcal{L}_{\bot}$ is to minimize the overlapping information carried by the projected features to properly exploit the unique characteristics of the two modalities.

\chapter{Implementation details}
\label{chapter:implementation}
The objective of this section is to describe more specifically the implementation components of the main parts of the network: RGB-/Depth-CNN, projection blocks and RNN. 
Showing the simplifications made, the implementation choices and the parameters used, thus allowing a repeatability of the performed experiments.
Nonetheless, it is important to keep in mind that the concepts described in section 3 are agnostic to the specific implementation that can be adapted to the task at hand.

\section{RGB/Depth Convolutional Neural Network}

The choice of the neural network was based on a compromise between network efficiency and the number of parameters. Since residual networks have become a standard choice, we deployed ResNet-18, the most compact representation proposed by He et al. ~\cite{he2016deep}.
Resnet18, as previously introduced, is a network that uses residual blocks to avoid the problem  of the vanishing of gradient. It is structured in 5 residual blocks for a total of 18 convolutional levels. It has about 40,000 parameters. An implementation of the network pre-trained on ImageNet is available here ~\cite{resnet-github}.

\section{Projection blocks}

The projection blocks transform a volumetric input into a vector of dimensions (1$\times$\texttt{pd}) through two convolutional layers and a global max pooling layer.
We designed the block in such a way that the first convolutional layer focuses on exploiting the spatial dimensions, width and height, of the input with \texttt{pd} filters of size (7$\times$7), while the second convolutional layer exploits its depth with \texttt{pd} filters of size (1$\times$1).
Finally, the global max pooling computes the maximum of each depth slice. This instantiation of the projection blocks has provided the best performances among those that we tried.
\section{Recurrent Neural Networks}
Keeping in mind that increasing the number of parameters also increase the difficulty of training the network itself. Remembering also that the dimensions of the datasets that we are used are considerably smaller compared to ImageNet.
The choice of the best implementation of RNN must be made by making a compromise between capacity and limited number of parameters.
The chosen RNN is called the \textbf{Gated Recurrent Unit (GRU)} ~\cite{gru}.
This network is considered to be a variation of long-short term memory (LSTM) ~\cite{lstm} and its effectiveness in dealing with long input sequences has been repeatedly shown ~\cite{chung2014}. Despite the two networks perform comparably, GRU requires 25\% less parameters than LSTM. In our experiments, we process the sequence of projected vectors with a single GRU layer with a number of memory neurons \texttt{mn} that is tweaked on the considered dataset. An implementation of GRU can be found in all the most popular deep learning libraries.

\chapter{Experiments}
\label{chapter:ex}
In this section we show the performance of the network using two RGB-D datasets representing objects: RGB-D Object Dataset~\cite{wrgbd} and JHUIT-50~\cite{jhuit50}.
The results of different experiments are reported, with different modality, in order to show the network in various aspects. 
Based on experimental evidences, we discuss strengths and weaknesses of our method, highlighting the most favourable and most problematic situations for the network. Finally, we compare our method to the state-of-the-art approaches on the object recognition and instance recognition task.

\section{Setup}

All the experiment reported in this section are performed on two popular RGB-D datasets, Washington RGB-D~\cite{wrgbd} and JHUIT-50~\cite{jhuit50}.
Since its introduction, Washington RGB-D dataset has become a reference point for the robotic community and it is the common choice for most of the existing methods in the field of RGB-D recognition.
This dataset contains 41,877 RGB-D images capturing 300 objects from 51 categories, spanning from fruit and vegetables to tools and containers.
For the evaluation, we follow the standard experimental protocol defined in \cite{bo2011}, where ten training/test split are defined in such a way that one object instance per class is left out of the training set. The reported results are the average accuracy over the different splits.
To show the efficiency of our method we decided to test it with a second dataset, always representing objects.\\
The second dataset, JHUIT-50~\cite{jhuit50}, contains 14,698 RGB-D images capturing 50 common workshop tools, such as clamps and screw drivers. This dataset presents few objects, but very similar to each others. 
For the evaluation, we follow the standard experimental protocol defined in~\cite{jhuit50}, where training data are collected from fixed viewing angles between the camera and the object while the test data are collected by freely moving the camera around the object. As stated in section~\ref{chapter:related_works}, we colorize the depth images of both datasets with surface normal encoding and adopt the pre-processing procedure used in~\cite{aakerberg2017}.

\begin{figure}[H]
\centering
\includegraphics[width=0.95\textwidth]{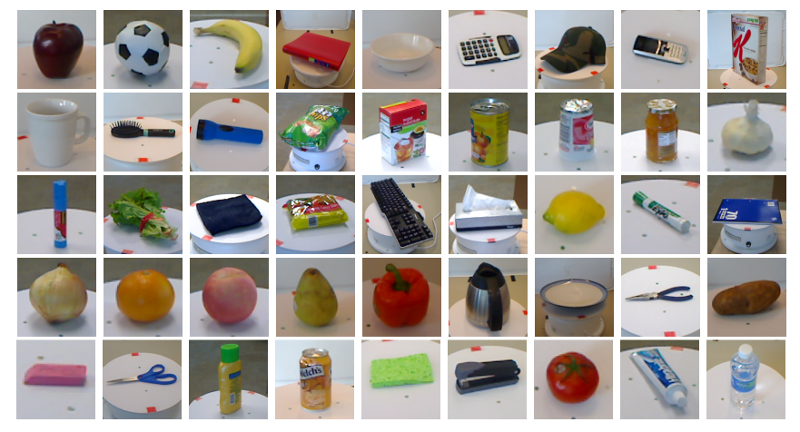} \quad \includegraphics[width=0.92\textwidth]{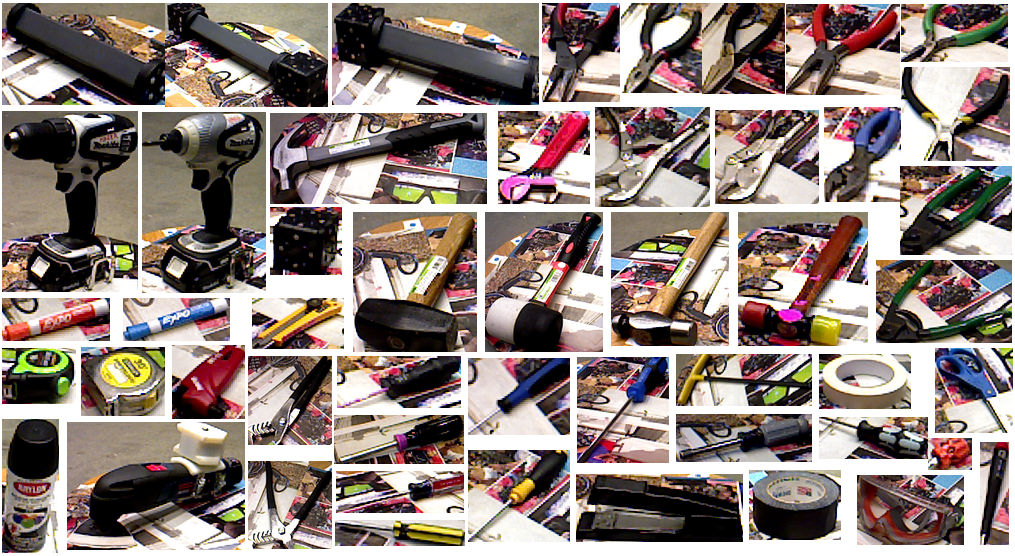}
\caption{Examples of RGB images belonging to the two datasets~\cite{wrgbd},~\cite{jhuit50}.}
\label{fig:dataset}
\end{figure}

\section{Training}

We trained our model using RMSprop optimizer ~\cite{rmsprop}. An implementation of this kind of optimizer can be found in all the most popular deep learning libraries, including TensorFlow~\cite{tensorflow}. 
We trained the network with the following parameters: batch size 64, learning rate 0.001, momentum  0.9, weight decay 0.0002 and max norm 4.
The weight $\lambda_i$  indicating the influence of the orthogonality loss for layer i, is set to 0.0001 for lower layers and decreases moving toward the output of the network. 
Through a grid search we found the best architecture parameters for the projection depth and the memory neurons, respectively $\texttt{pd}=512$ and $\texttt{mn}=50$.
The weights of the two ResNet-18 used as the RGB- and Depth-CNN are initialized with values obtained by pre-training the networks on ImageNet. 
Currently we have an average of over five hundred images per node. 
The rest of the network is initialized with Xavier initialization method \cite{xavier} in a multi-start fashion where the network is initialized multiple times and, after one epoch, only the most promising model continues the training.\\
The same kind of research is done for JHUIT-50, but the small size of the dataset made the training of the network difficult, producing low values of accuracy.
For this reason, we compensate  the small training set of about $7,000$ images with simple data augmentation techniques: scaling, horizontal and vertical flip, $90$ degree rotation.
Obtaining a training set 7 times larger than the previous one that allowed us to better train our network, increasing the final performances

\section{Results}

This section reports the results of different experiments to show the capabilities of the network, highlight the innovative aspects and provide a direct comparison with the other existing methods.
The first set of experiments aims to show the effects of the assumptions and implementation choices made on the network: we isolate the contribution on the two main components of RCFusion: multi-level feature extraction and orthogonality loss.
Then, we provide some hints to understand the behavior of the method by analyzing its performance on specific object categories.
Finally the performances of our architecture are compared with other existing methods that we have found in the literature.
For sake of compactness, the firsts two set of experiments are performed on the RGB-D Object Dataset, while the benchmark includes also the results on JHUIT-50 to demonstrate the robustness of our method against changes in datasets.

\subsection{Ablation study}

From the results in table~\ref{tab:ablation}, we can see how the network performances are influenced by the level of abstraction of the features that feed the RNN. We progressively consider features extracted from lower layers of the two ResNet-18. 
More precisely, we start from the output of the last residual block (output) and go backward until to consider also the first residual block (\textit{res1}). 
We repeat the same experiments by removing $\mathcal{L_{\bot}}$ from the loss function. 
The results in table~\ref{tab:ablation} show that the sole inclusion of lower level features can improve the accuracy of the network. We can see as all the configurations of the network,from \textit{res1-5} to \textit{res5}, achieve  a better accuracy respect the $93.9\%$ of the case where consider only the output. 
When including $\mathcal{L_{\bot}}$ in the optimization process, the accuracy can further increase, in the case of best performance we arrive to achieve $94.4\%$. It is worth pointing out that the weight $\lambda_i$ for layers higher than the forth residual block is set to zero, thus motivating the missing values in table \ref{tab:ablation}. 
The two configurations \textit{res2-5} and \textit{res3-5} always have higher values than the others.
We can infer that the features from a level of abstraction very low, as the first residual block, don’t help the training of the network. At the same time the features of a high level, as \textit{res5}, are not enough for the training of RNN.
\begin{table}
\small
\begin{center}
\begin{tabular}{|l|c|c|c|c|c|c|}
\hline
\multicolumn{7}{|c|}{\textbf{Ablation study}} \\ \cline{1-7}
\hline
Loss & res1-5 & res2-5 & res3-5 & res4-5 & res5 only & output \\
\hline\hline
w/ $\mathcal{L_{\bot}}$ &
 94.0$\pm$1.1 & 94.3$\pm$1.0 & 94.4$\pm$1.4 & 94.0$\pm$1.1 & - & - \\
w/o $\mathcal{L_{\bot}}$ &
 93.9$\pm$1.3 & 94.2$\pm$1.7 & 94.2$\pm$1.1 & 94.0$\pm$1.4 & 94.0$\pm$1.7 & 93.9$\pm$1.0 \\
\hline
\end{tabular}
\end{center}
\caption{Accuracy (\%) of different configurations of our method: $resA-B$ indicates that the features are extracted from residual block $A$ until $B$ of ResNet-18 and \textit{output} indicates the output of the final residual block; w/(o) $\mathcal{L_{\bot}}$ indicates that the orthogonality loss is included (or not) in the optimization process. }
\label{tab:ablation}
\end{table}

\subsection{Analysis}

\begin{figure*}[h]
\centering
\includegraphics[width=0.97\textwidth]{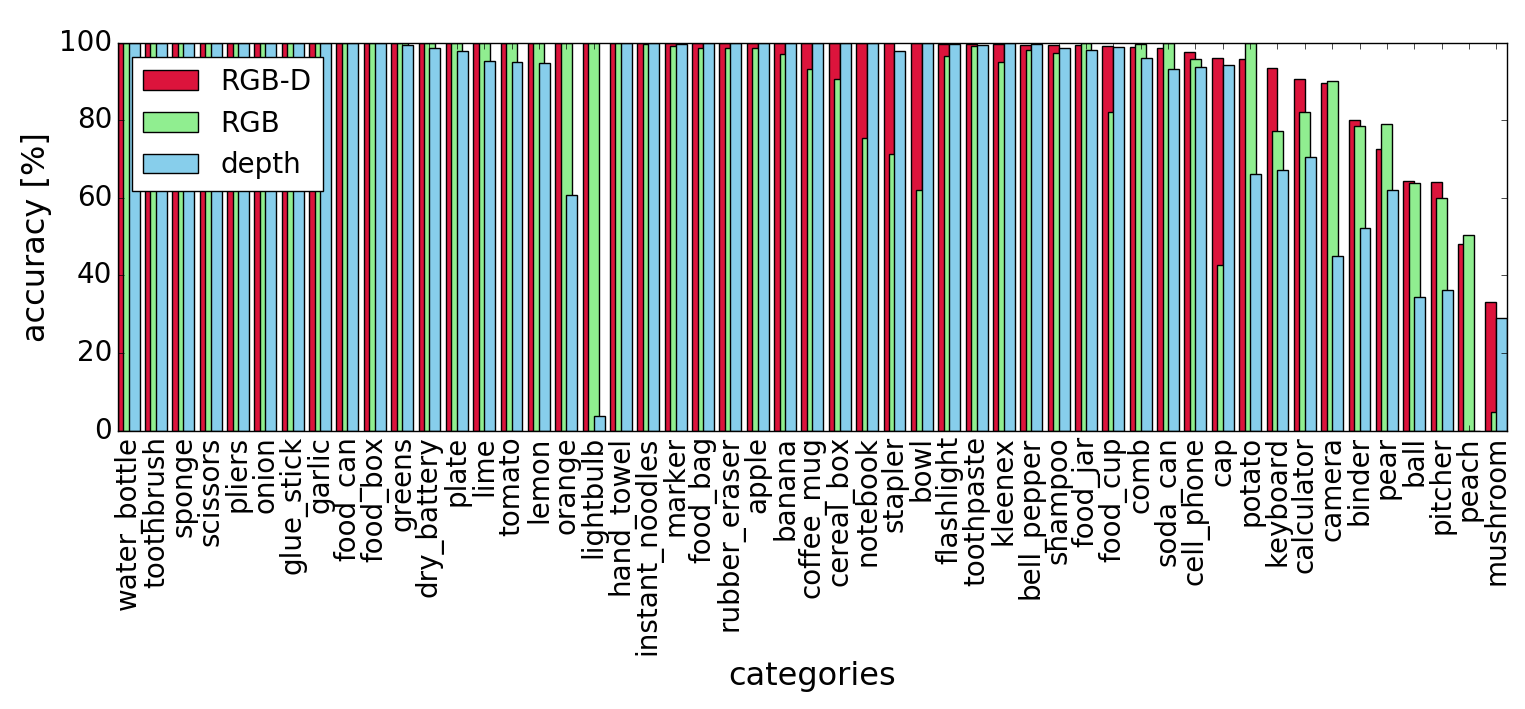}
\caption{Per class accuracy ($\%$) of RCFusion on RGB-D Object Dataset~\cite{wrgbd}.}
\label{fig:per_class}
\end{figure*}

In this section it is shown a more detailed analysis of the results obtained from the network.
Showing the strengths and weaknesses of the network. Trying to identify the conditions where effectively the network succeeds in introducing an improvement over the individual modalities.
In order to better understand the performance of RCFusion, we report the per-class accuracy on Washington RGB-D in figure ~\ref{fig:per_class}. 
The first observation we can make from the graph is that the multi-modal approach either matches or improves over the results on the single modalities for almost all categories. 
For categories like \textit{"lightbulb"}, \textit{"orange"} or \textit{"bowl"}, where the accuracy on one modality is very low, RCFusion learns to rely on the other modality.
There are also situations in which the multi-modal produces a lower result than a single mode, like for \textit{"pear"} and \textit{"potato"}.
This happens in the specific case in which  an object class is confused with the same classes in both RGB and depth modalities, This highlights a weakness of the method that will be the subject of future investigations.
Instead, when an object class is confused with distinct classes in the individual modalities, like for \textit{"keyboard"} and \textit{"calculator"}, the RGB-D modality can perform better. \\
It is possible to see the multimodal's contribution compared to the single modalities also through observing the t-SNE graph of the final features. It is evident how the RGB-D features of each category are more compact clustered than the single modalities.

\begin{figure*}[h]
\centering
\includegraphics[width=0.48\textwidth]{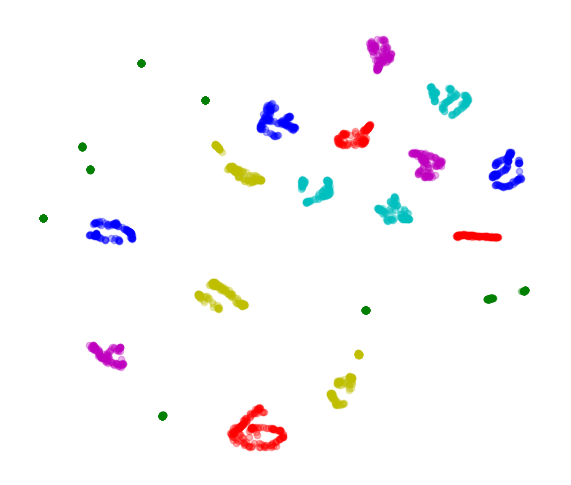}
\includegraphics[width=0.50\textwidth]{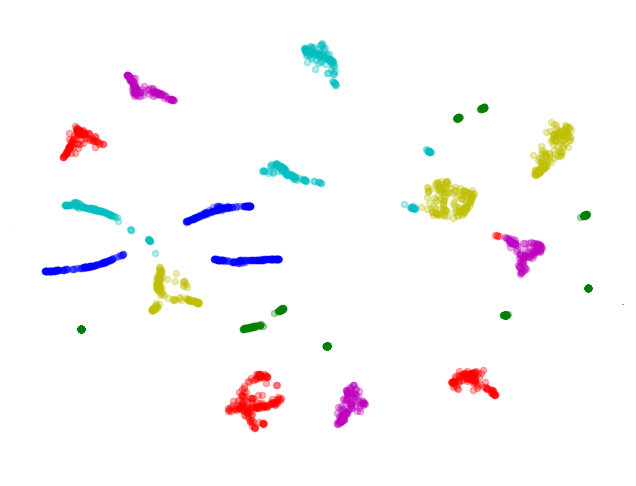}
\includegraphics[width=0.71\textwidth]{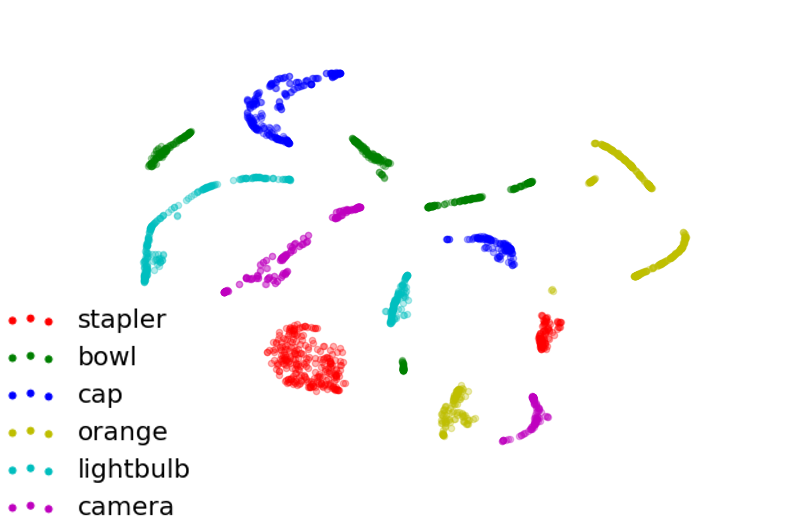}
\caption{t-SNE visualization of the final features obtained for RGB, depth and RGB-D modalities.}
\label{fig:tsne}
\end{figure*}

\subsection{Benchmark}
We benchmark RCFusion on RGB-D Object Dataset and JHUIT-50 against other methods in the literature. \\
The results in table \ref{tab:wrgbd} for the individual RGB and depth modality show that ResNet-18 is a good trade-off between limited number of parameters and high accuracy. 
The accuracy that we have obtained on the RGB modality is second only to \cite{fisher}, where they use a VGG network~\cite{vgg} that introduces considerably more parameters than ResNet-18.
For the depth modality, ResNet-18 achieves higher accuracy than all the competing methods. 
Regarding the final multi-modal accuracy, our method clearly outperforms all the competing approaches. It is worth noticing that, when we extract the features only from the output of the last residual layer (RCFUsion -- output), we achieve a value comparably to the second best method \cite{fisher}. 
However, when including lower level features, our method gains a $+0.6\%$ over \cite{fisher}. 
This highlights, once again, the importance of multi-level feature extraction and orthogonality loss as the main contributions of this work.\\
Table~\ref{tab:jhuit} shows the results on JHUIT-50. For the individual modalities, ResNet-18 shows again a compelling performance. In the multi-modal RGB-D classification, our method clearly outperforms all the competing approaches with a margin of $2\%$ on~\cite{deco}. \\
In summary, RCFusion establishes new state-of-the-art results on the two most popular datasets for RGB-D object recognition, demonstrating its robustness against changes in the dataset and the task.

\begin{table}[h]
\small
\begin{center}
\begin{tabular}{|l|c|c|c|}
\hline
\multicolumn{4}{|c|}{\textbf{RGB-D Object Dataset}} \\ \cline{1-4}
\hline
Method & RGB & Depth & RGB-D \\
\hline\hline
LMMMDL~\cite{wang2015} &
 74.6$\pm$2.9 & 75.5.8$\pm$2.7 & 86.9$\pm$2.6\\
FusionNet~\cite{eitel2015} &
 84.1$\pm$2.7 & 83.8$\pm$2.7 & 91.3$\pm$1.4 \\
 CNN w/ FV~\cite{li2015} &
 {\color{red}\textbf{90.8}}$\pm$1.6& 81.8$\pm$2.4 & {\color{blue}\textbf{93.8}}$\pm$0.9 \\
 DepthNet~\cite{depthnet} &
 88.4$\pm$1.8 & 83.8$\pm$2.0 & 92.2$\pm$1.3 \\
 CIMDL~\cite{cimdl} &
 87.3$\pm$1.6 &  {\color{blue}\textbf{84.2}}$\pm$1.7 & 92.4$\pm$1.8 \\
 FusionNet enhenced~\cite{aakerberg2017} &
 {\color{blue}\textbf{89.5}}$\pm$1.9 & {\color{blue}\textbf{84.5}}$\pm$2.9 & {\color{blue}\textbf{93.5}}$\pm$1.1 \\
 DECO~\cite{deco} &
 {\color{blue}\textbf{89.5}}$\pm$1.6 & 84.0$\pm$2.3 & {\color{blue}\textbf{93.6}}$\pm$0.9 \\\hline
 RCFusion -- output &
 {\color{blue}\textbf{89.6}}$\pm$2.2 & {\color{red}\textbf{85.9}}$\pm$2.7 & {\color{blue}\textbf{93.9}}$\pm$1.0 \\
  RCFusion -- res3-5 &
 {\color{blue}\textbf{89.6}}$\pm$2.2 & {\color{red}\textbf{85.9}}$\pm$2.7 & {\color{red}\textbf{94.4}}$\pm$1.4 \\
\hline
\end{tabular}
\end{center}
\caption{Accuracy (\%) of several methods for object recognition on RGB-D Object Dataset~\cite{wrgbd}. Red: highest result; blue: other considerable results.}
\label{tab:wrgbd}
\end{table}

\begin{table}[h]
\small
\begin{center}
\begin{tabular}{|l|c|c|c|}
\hline
\multicolumn{4}{|c|}{\textbf{JHUIT-50}} \\ \cline{1-4}
\hline
Method & RGB & Depth & RGB-D \\
\hline\hline
 DepthNet~\cite{depthnet} &
 88.0 & 55.0 & 90.3 \\
 FusionNet enhenced~\cite{aakerberg2017} &
 {\color{blue}\textbf{94.7}} & 56.0 & {\color{blue}\textbf{95.3}} \\
 DECO~\cite{deco} &
 {\color{blue}\textbf{94.7}} & {\color{red}\textbf{61.8}} & {\color{blue}\textbf{95.7}} \\\hline
 RCFusion (ours) &
 {\color{red}\textbf{95.1}} & {\color{blue}\textbf{59.8}} & {\color{red}\textbf{97.7}} \\
\hline
\end{tabular}
\end{center}
\caption{Accuracy (\%) of several methods for object recognition on JHUIT-50~\cite{jhuit50}. Red: highest result; blue: other considerable results.}
\label{tab:jhuit}
\end{table}

\chapter{Conclusion and future work}
\label{chapter:conclusion}
In the previous chapters we presented a deep multi-modal network for RGB-D object recognition which takes the name of Recurrent Convolutional Fusion.
The network developed in this thesis introduces the innovative assumption that adding lower level features can help generate more discriminative RGB-D features.
Our method uses two streams of convolutional networks to extract RGB and depth features from multiple levels of abstraction. These features are then concatenated and sequentially fed to an RNN to obtain a compact RGB-D feature that is used by a softmax classifier for the final classification. An orthogonality loss is also adopted to encourage the two streams to learn complementary information.\\
To evaluate the performance of our network we used two datasets: RGB-D Object Dataset and JHUIT-50.
We have obtained, for both datasets, a result that exceeds the current state of the art.
A detailed analysis of the strengths of our approach is performed. Analyzing the disadvantageous cases that the network can not manage and that represent a starting point for future studies.\\
The project developed in this thesis has as main objective to show the validity of the assumption made at the beginning.
Obviously the performances obtained are also dependent on other factors such as the types of network used and type of colorization chosen for the depth.
It must be interesting to evaluate the performances using different types of CNNs such as Resnet50, which re-presents a network based on residual blocks as in Resnet18.
A possible future development could also be that of using other types of RNN as well: vanilla RNN, LSTM.\\
From the results shown in the previous paragraphs we observe how, very often, the information extracted from the depth are not as useful as the RGB information.
This underlines how the performance of this approach can improve with new techniques, more efficient than surface normal ~\cite{bo2013}.
The choice of surface is made to make the comparison with other methods more fair.
We want to underline that the results achieved are mainly due to the innovative assumption at the base of the network.
Choosing more efficient methods of colorization could introduce doubts on the effective innovation and validity of this new approach, making the comparison with the other methods not very useful.
Once we have verified our original ideas, a possible development could be trying different methods of colorizing the depth, such as Deco ~\cite{deco}, and analyzing how performance changes.\\
Due to their implementation-agnostic nature, the main concepts presented in this work can be adapted to different tasks. In fact, the results obtained on object categorization encourage further research to extend this approach to higher level tasks, such as object detection and semantic segmentation.

\newpage

\addcontentsline{toc}{chapter}{\bibname}

\printbibliography
\end{document}